\documentclass{bmvc2k}

\usepackage{times}
\usepackage{epsfig}
\usepackage{graphicx}
\usepackage{amsmath}
\usepackage{amssymb}

\usepackage{booktabs}
\usepackage{multirow}
\usepackage{tabularx}
\usepackage{wrapfig}
\usepackage{soul}
\usepackage{bbm}
\usepackage{sidecap}
\newcolumntype{P}{>{\centering\arraybackslash}X}

\usepackage{enumitem}
\usepackage{multirow}
\usepackage{wrapfig}
\usepackage{colortbl}
\usepackage[normalem]{ulem}

\setlist[itemize]{align=parleft,left=0pt}

\definecolor{azure(colorwheel)}{rgb}{0.0, 0.5, 1.0}
\definecolor{nicegreen}{rgb}{0.0, 0.7, 0.1}
\definecolor{yw}{rgb}{0.784, 0.003, 0.313}
\definecolor{ywg}{rgb}{0.9960, 0.8984, 0.5859}
\definecolor{jy}{rgb}{0.58, 0, 0.827}
\definecolor{CuGray}{gray}{0.9}

\definecolor{rev}{rgb}{0.784, 0.003, 0.313}
\definecolor{pink}{cmyk}{0, 0.7808, 0.4429, 0.1412}
\definecolor{amethyst}{rgb}{0.6, 0.4, 0.8}
\definecolor{black}{rgb}{0.0, 0.0, 0.0}
\definecolor{tb3_yellow}{rgb}{0.996, 1.0, 0.6}
\definecolor{tb3_orange}{rgb}{0.980, 0.8, 0.604}
\definecolor{tb3_red}{rgb}{0.972, 0.6, 0.6}

\definecolor{blue}{rgb}{0.0, 0.0, 0.4}

\usepackage[font={footnotesize,color=blue}]{caption}

\newcolumntype{g}{>{\columncolor{CuGray}}c}
\newcolumntype{z}{>{\columncolor{CuGray}}l}

\renewcommand{\paragraph}[1]{\noindent\textbf{#1.}\,\,}

\newcommand{\ccy}[1]{\cellcolor{tb3_yellow}{#1}}
\newcommand{\cco}[1]{\cellcolor{tb3_orange}{#1}}
\newcommand{\ccr}[1]{\cellcolor{tb3_red}{#1}}
\newcommand\inv[1]{#1\raisebox{1.25ex}{$\scriptscriptstyle-\!1$}}
\usepackage{xspace}

\def\onedot{.\@\xspace}
\def\eg{\emph{e.g}\onedot} 
\def\ie{\emph{i.e}\onedot}

\def\etal{\emph{et al}\onedot}

\newcommand{\Sref}[1]{Sec.~\ref{#1}}

\newcommand{\Fref}[1]{Fig.~\ref{#1}}
\newcommand{\Tref}[1]{Table~\ref{#1}}

\def\DeMR{\emph{DeMR}}

\newcommand{\be}{\begin{eqnarray}}
\newcommand{\ee}{\end{eqnarray}}
\newcommand{\bee}{\begin{eqnarray*}}
\newcommand{\eee}{\end{eqnarray*}}

\newcommand{\matrixb}{\left[ \begin{array}}
\newcommand{\matrixe}{\end{array} \right]}

\newcommand{\para}[1]{\vspace{1mm}\paragraph{#1}}

\title{Unified 3D Mesh Recovery of Humans and Animals by Learning Animal Exercise}

\addauthor{Kim Youwang}{youwang.kim@postech.ac.kr}{1}{}
\addauthor{Kim Ji-Yeon}{jyeonkim@postech.ac.kr}{2}{}
\addauthor{Kyungdon Joo}{kyungdon@unist.ac.kr}{4}{,5}
\addauthor{Tae-Hyun Oh}{taehyun@postech.ac.kr}{1}{,2,3}
\addinstitution{
 Dept. of Electrical Eng.\\
 POSTECH, Korea
}
\addinstitution{
 Dept. of Convergence IT Eng.\\
 POSTECH, Korea
}
\addinstitution{
 Grad. School of Artificial Intelligence\\
 POSTECH, Korea
}
\addinstitution{
 Artificial Intelligence Grad. School\\
 UNIST, Korea
}
\addinstitution{
 Dept. of Comp. Science and Eng.\\
 UNIST, Korea
}

\runninghead{Youwang et al.}{Unified 3D Mesh Recovery of Humans and Animals}

\begin{document}

\maketitle
\vspace{-5mm}
\begin{abstract}
    We propose an end-to-end unified 3D mesh recovery of humans and quadruped animals trained in a weakly-supervised way.  
    Unlike recent work focusing on a single target class only, we aim to recover 3D mesh of broader classes with a single multi-task model.
    However, there exists no dataset that can directly enable multi-task learning due to the absence of both human and animal annotations for a single object, \eg, a human image does not have animal pose annotations; thus, we have to devise a new way to exploit heterogeneous datasets.
    To make the unstable disjoint multi-task learning jointly trainable, we propose to exploit the morphological similarity between humans and animals, motivated by \emph{animal exercise} where humans imitate animal poses.
    We realize the morphological similarity by semantic correspondences, called sub-keypoint, which enables joint training of human and animal mesh regression branches.
    Besides, we propose class-sensitive regularization methods to avoid a mean-shape bias and to improve the distinctiveness across multi-classes.
    Our method performs favorably against recent uni-modal models on various human and animal datasets while being far more compact.
    \vspace{-3mm}
 
\end{abstract}

\vspace{-4mm}
\section{Introduction}
\vspace{-1mm}

\begin{wrapfigure}{R}{0.21\linewidth}
\centering
    \includegraphics[width=1\linewidth]{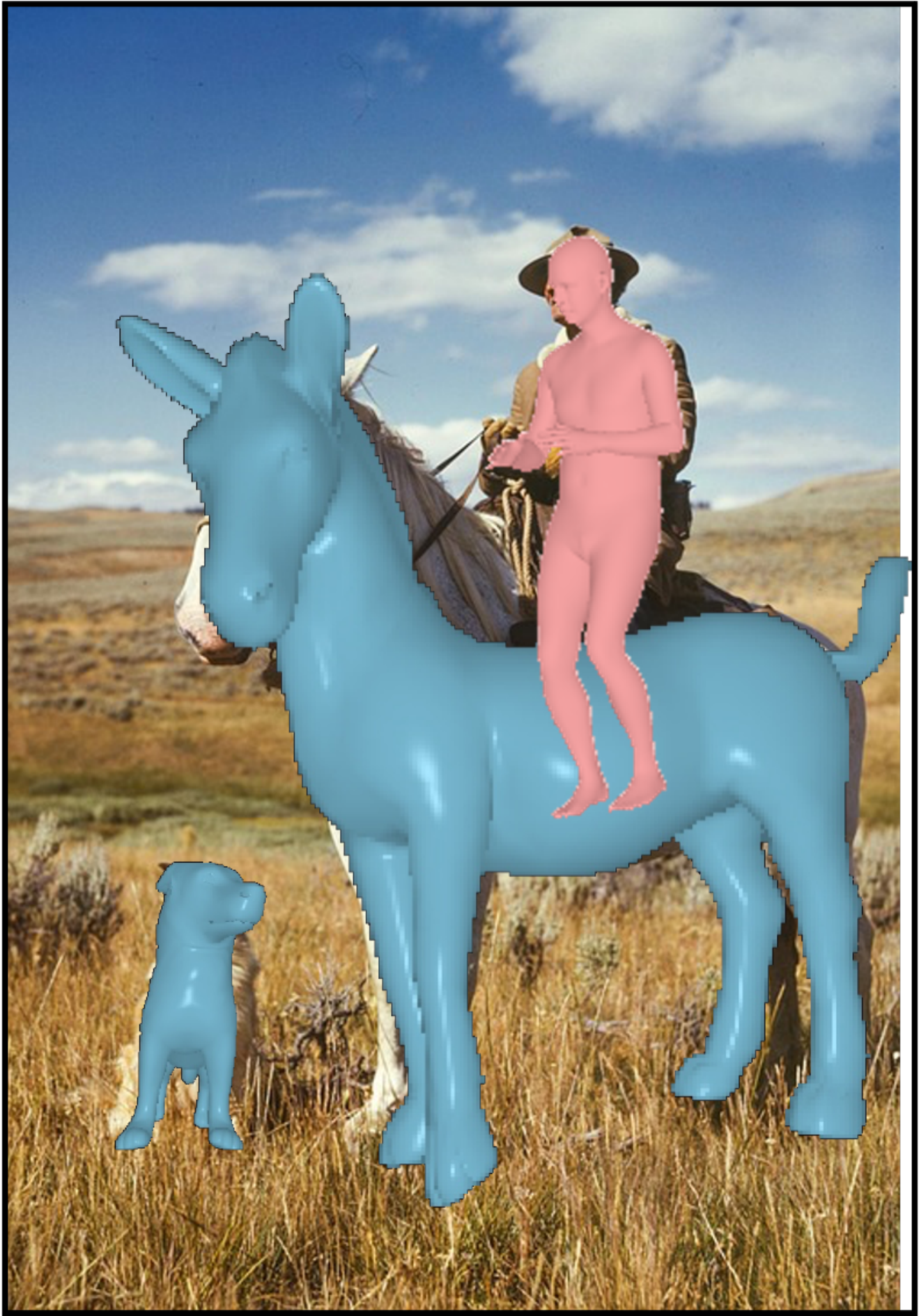}\vspace{2mm}
    \caption{\DeMR\, is a unified model that reconstructs meshes of humans and animals.\protect\footnotemark
    }
    \vspace{-4mm}
    \label{fig:teaser}
\end{wrapfigure}


Considering the fact that countless species of animals are present in nature, it is challenging to develop a machine that understands the behavior of general animals. To resolve this challenge, the humans' capability of imitating animals, \ie,~\emph{animal exercise}, can be a crucial clue to build such a machine.  
Concretely, humans can express and mimic animals by fully utilizing their body parts similar to those of animals during animal exercise, which is capable 
thanks to the morphological similarity between humans and animals.
Inspired by this morphological similarity,
we tackle a unified multi-task model that estimates the 3D volumetric meshes of deformable objects from a single image (see \Fref{fig:teaser}), including humans and a few quadruped animals, called \textit{Deformable Objects Mesh Recovery}, \DeMR.

A major challenge to develop such a unified model is the requirement of heterogeneous datasets of humans and animals
that have incompatible label formats, 
\ie, disjoint label set setup~\cite{li2017learning,kim2018disjoint}.
In other words, human images do not have animal keypoint annotations or \emph{vice versa}.
It is known that
directly deploying multi-task learning with disjoint label sets by
a na\"ive approach yields
performance degradation compared to uni-modal models~\cite{li2017learning,kim2018disjoint}.
We tackle this challenge by leveraging 
a few sub-keypoints, which is a subset of full body keypoints morphologically shared across heterogeneous classes, \eg, \{left wrist: left front paw\}.
We enforce the model to at least fit morphologically corresponding keypoints, such that it can implicitly learn such similarity among different classes.
This enables stable multi-task learning by jointly computing the losses of each 
multi-task branch even without compatible annotations across all the classes. 
Lastly, despite this mutually beneficial development, since our model mainly focuses on learning
shared knowledge, it could be biased to predict mean-like shapes for different species. 
We mitigate it by introducing class-sensitive regularizations: 
class-selective shape prior loss and class-specific batch normalization layers.

\footnotetext{We performed \DeMR\, for each cropped image and then composited them for visualization.}

On top of these developments, our proposed model can reconstruct realistic and distinctive 3D shapes of different quadruped animals as well as those of humans.
As a favorable by-product, our training scheme can mitigate data scarcity issues that exist in the animal keypoint annotated datasets.
Besides, our unified model benefits shared knowledge of multiple heterogeneous classes, whereby the model can preserve comparable or better performance than each of the uni-modal models without increasing the model size.
We show 
our model’s 3D reconstruction performance
with 
competing methods in human and animal domains~\cite{hmrKanazawa17,biggs2020wldo}.
We summarize our main contributions as follows: \vspace{-1mm}
\begin{itemize}
    \item We present {\em DeMR}, a unified neural regression model that can directly reconstruct 3D pose and shape of various classes (humans and quadruped animals) from a single image.\vspace{-2mm}
    \item We suggest
    a data-efficient way to deal with cumbersome training with 
    human and animal datasets with incompatible labels. We tackle it by introducing joint multi-task learning using sub-keypoint
    that leverages morphological similarity among heterogeneous classes.\vspace{-2mm}
    \item We propose novel
    methods to tackle the challenges due to
    the domain gaps among heterogeneous classes:
    class-selective shape prior loss and 
    class-specific normalization layer.
    
\end{itemize}
\vspace{-5mm}


\section{Related Work}
\vspace{-2mm}
Our task is related to the multi-class
3D full-body shape and pose estimation.
The 3D volumetric recovery has been mainly studied using
parametric body models, such as SMPL~\cite{SMPL:2015} and SMAL~\cite{Zuffi:CVPR:2017}, called model-based approaches. 
We briefly review 
these lines of researches.

\para{Monocular 3D Human Reconstruction} 
Recent advances in body models and model-based 3D human reconstruction have enabled accurate 3D human mesh reconstruction
from a single image.
SMPL provides a compact representation of volumetric 3D body mesh with interpretable parameterization of shape, pose, and camera parameters.
SMPLify~\cite{bogo2016smpl} introduced the first fully automatic approach to fit the SMPL model onto an image via an optimization pipeline.
%
Later, HMR~\cite{hmrKanazawa17} introduced a learning-based SMPL parameter regressor using Convolutional Neural Network (CNN). 
HMR directly regresses mesh parameters using extracted body features from CNN.
Compared to the optimization-based method, the CNN-based method enables efficient estimation and opens versatile applications.
%
Recently, numerous other approaches, \eg, optimization in-the-loop, self attention-based vertex regression~\cite{kolotouros2019spin,lin2021endtoend}, have achieved significant advances by focusing only on humans.
%
Our \DeMR\ leverages HMR as a backbone to obtain shared body features across heterogeneous classes, extending beyond the human class.

\para{Monocular 3D Animal Reconstruction} 
Analogous to the human parametric model, the 3D animal body model, \ie, SMAL~\cite{Zuffi:CVPR:2017}, has boosted the development of model-based 3D animal reconstruction.
%
%
Optimization-based approaches~\cite{Zuffi2018LionsAT,biggs2018creatures} 
and learning-based methods~\cite{9010937,biggs2020wldo} were suggested to infer 3D animal body meshes
onto a single image.
However, despite these advances, 3D animal reconstruction has been less spotlighted than that of humans.
The reason stems from insufficient 
animal shape
annotations and limited animal class coverage for 3D motion capture datasets.
Such limitations have narrowed down the research scope of the previous works to only cover one or few animal classes.
%
We leverage the morphological similarity across different animal and human classes.
This enables a unified model to be well-supervised in data-scarce regimes of a few classes, which deals with the limitation of the prior works.
In contrast to recent works
that focus on 3D mesh reconstruction of either only a single object class at a time or a similar class of animals, \ie, uni-modal model, 
our method is the first learning-based method that can directly estimate 3D meshes of multiple heterogeneous classes, including humans and quadruped animals, and do not require any test time annotations.

\para{Pose Estimation of Multiple Animal Classes} 
Aside from 3D reconstruction tasks of a single object class, 
there have been a few attempts to build a unified model of multiple object classes but with 2D or 2.5D pose representations.
%
Cao~\etal\cite{Cao2019CrossDomainAF} showed physical similarity among humans and animals can be learned by estimating 2D keypoints with cross-domain adaptation.
With 2.5D representation, 
recent work~\cite{densepose-chimp} showed that physically analogous 2.5D pose representation of humans can be extended to primates, \eg, gorilla and chimpanzee.
%
Recently, functional maps~\cite{funcmaps} opened up an opportunity to extend the 2.5D pose representation of humans to broader animal classes~\cite{NEURIPS2020_c81e728d}. 
Although these works have limited scopes of pose representations, they have proved the existence of morphological similarity among heterogeneous classes. 
We leverage this insight to build our unified model but tackle a more challenging regime, \ie, 3D pose and shape with a unified model. 
\vspace{-4mm}

\section{\DeMR: Deformable Objects Mesh Recovery}
\vspace{-2mm}
%
In this section,
we first define the term, \textit{deformable objects}, and review 
body representations for deformable objects
in \Sref{sec:preliminary}. 
Next, we present the overall structure
of our method 
in \Sref{sec:overview}.
%
Then, we propose novel methods to enable 
learning \textit{morphological similarity} and class-specific regularizations
in \Sref{sec:morph}.

\vspace{-4mm}


\subsection{Preliminary} \label{sec:preliminary}
\vspace{-2mm}
We consider
four deformable object classes that frequently appear in the wild nature with distinct body poses and shapes, and 
the classes 
that have a reasonable amount of data available for training and evaluation. 
Accordingly, we consider \textit{Human, Dog, Horse} and \textit{Cow} as our target deformable object classes.

\para{Body Representation of Deformable Objects} 
We adopt parametric 3D body models to model 3D bodies of deformable objects that we consider.
SMPL~\cite{SMPL:2015} represents a human body with the 3D mesh vertices $M_h\in{\mathbb R}^{6890\times3}$ by a differentiable function $\small{{\mathcal M}_h(\theta_h,\beta_h)}$ that takes the human pose parameters $\theta_h\in{\mathbb R}^{24\times3}$ 
in angle-axis representation 
and the human shape parameters $\beta_h\in{\mathbb R}^{10}$ as input. 
Correspondingly, as the quadruped animal model, we use SMAL~\cite{Zuffi:CVPR:2017} that represents the 3D mesh vertices $M_a\in{\mathbb R}^{3889\times3}$ by ${\mathcal M}_a(\theta_a,\beta_a)$ that takes animal pose parameters $\theta_a\in{\mathbb R}^{35\times3}$ and animal shape parameters $\beta_a\in{\mathbb R}^{20}$ as input.
%
%
\vspace{-2mm}


\subsection{\DeMR's Architecture and Loss Functions} \label{sec:overview}
\begin{figure*}[t]
\centering
   \includegraphics[width=0.85\linewidth]{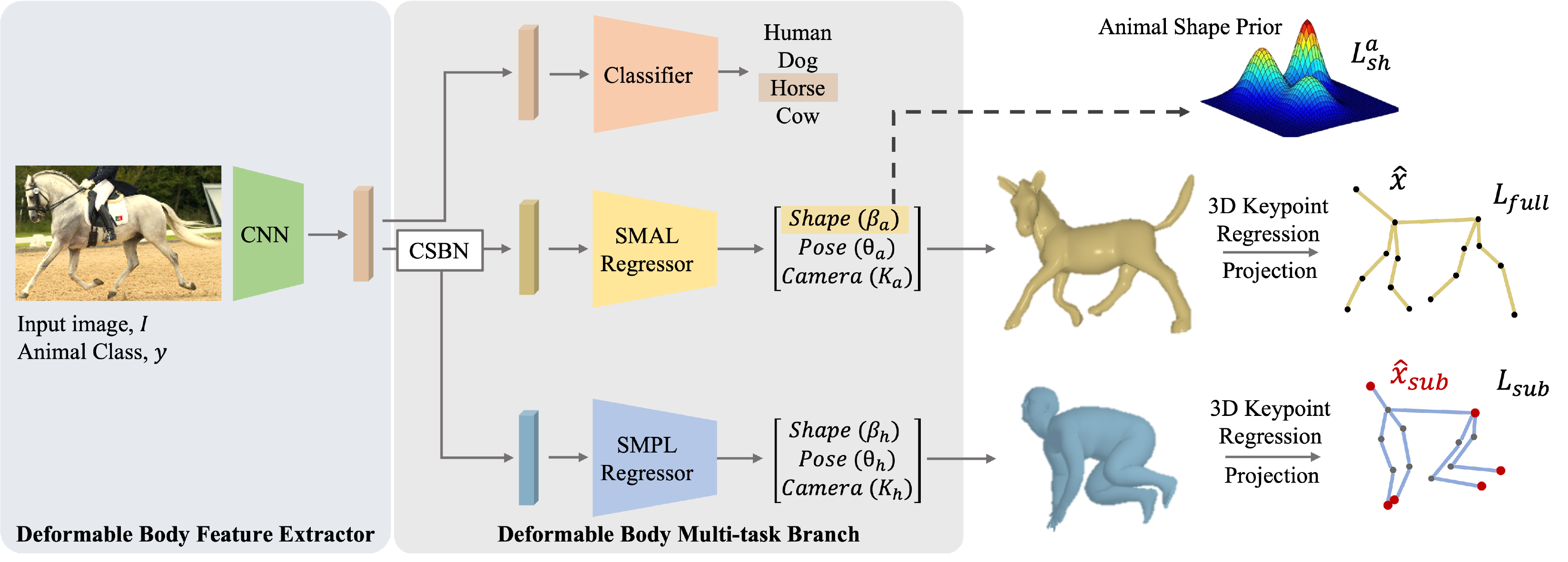}
   \vspace{3mm}
   \caption{{\textbf{\DeMR's Architecture}. During training, given an input image, the model regresses SMPL and SMAL mesh parameters, and class probability in a multi-task manner. At test time, \DeMR\, can (i) \textit{regress} both SMPL and SMAL parameters regardless of input object class, and (ii) \textit{select} which parameter it should use to construct appropriate mesh, according to the classification result.}}
   \vspace{-4.5mm}
\label{fig:network}
\end{figure*}
%
\vspace{-2mm}
The overall architecture of \DeMR\ mainly consists of two parts: \textit{Deformable body feature extractor} and \textit{Deformable body multi-task branch} (refer to \Fref{fig:network}). 
%

\para{Deformable Body Feature Extractor} 
It is a shared CNN that takes a single center-cropped deformable object image as an input and extracts a single feature vector $\mathbf{f}\in\mathbb{R}^{2048}$. 
%
We use the same backbone network
with the previous SMPL
regression works~\cite{hmrKanazawa17, kolotouros2019spin}. 
We guide this shared network
to learn shared body morphology across heterogeneous classes. 

\para{Deformable Body Multi-task Branch} 
The multi-task branches take the extracted feature vector as input and perform parameter estimation for 3D body reconstruction 
and classification in a multi-task manner.   
%
Each task is conducted via respective fully connected layers. 
The extracted feature vector encodes 
appearance information so that the model can perform object classification. 
This classification
allows the model to select which mesh parameters (or branch) to be used at 
test time.
%
The other multi-task branches perform class-specific mesh parameter estimation from a shared feature. 
Note that, 
since multi-class datasets are mixed in a batch during training, a
statistical gap
between humans and animals exists.
%
Thus, we propose class-sensitive regularizations to mitigate the gap and facilitate the multi-task branches to learn the class-specific residual information, described later. 

\para{Loss Function}
Our model is trained with 2D keypoint annotations mainly, and silhouette masks are used if available, \ie, no 3D annotation is used.
Our total loss $L$ is defined as:
\begin{equation}
     \small{L =\alpha (\lambda_{\scriptscriptstyle{f}}^{h}L_{\scriptscriptstyle{full} }^{h}{+}\lambda_{\scriptscriptstyle{s}}L_{\scriptscriptstyle{sub}}^{h})+(1{-}\alpha)(\lambda_{\scriptscriptstyle{f}}^{a}L_{\scriptscriptstyle{full}}^{a}{+}\lambda_{\scriptscriptstyle{s}}L_{\scriptscriptstyle{sub}}^{a})} 
    + \lambda_{\scriptscriptstyle{sil}}L_{\scriptscriptstyle{sil}} + \lambda_{\scriptscriptstyle{sh}}^{a}L_{\scriptscriptstyle{sh}}^{a},
\end{equation}
\noindent where $L^{*}_{\scriptscriptstyle{full}}$
are full-keypoint reconstruction loss terms, $L_{\scriptscriptstyle{sub}}^{*}$
are sub-keypoint reconstruction loss terms, $L_{\scriptscriptstyle{sil}}$ is a silhouette reconstruction loss term, and $L_{\scriptscriptstyle{sh}}^{a}$ is a class-selective shape prior loss. 
%
Besides, $\{\lambda_{\,*}\}$
denotes the balance parameters 
for each loss term, 
and $\alpha$ 
a human/animal indicator; \ie, $\alpha=1$ 
for human data, $\alpha=0$ otherwise. 
Details about hyper-parameters 
can be found 
in the supplementary material.
%


Following recent works~\cite{bogo2016smpl,hmrKanazawa17,kolotouros2019spin}, we use 
the full 2D keypoint reconstruction loss as the primary
training loss.
Once we reconstruct 
SMPL and SMAL meshes
from the estimated mesh parameters, ${\mathbf \theta_{\scriptscriptstyle h}}$, $\beta_{\scriptscriptstyle h}$, ${\mathbf \theta_{\scriptscriptstyle a}}$ and $\beta_{\scriptscriptstyle a}$ and the respective camera parameters, $K_{\scriptscriptstyle h}$ and $K_{\scriptscriptstyle a}$,
we can regress 3D keypoints
${\mathbf J}_{\scriptscriptstyle h}$ and ${\mathbf J}_{\scriptscriptstyle a}$ by the linear combination of each subset of the mesh vertices.
By orthographic projection of the estimated 3D keypoints, as ${\hat{\mathbf{x}}_{\scriptscriptstyle{h}}}=\Pi({\mathbf J}_{\scriptscriptstyle h}; K_{\scriptscriptstyle h})$ and ${\hat{\mathbf{x}}_{\scriptscriptstyle a}}=\Pi({\mathbf J}_{\scriptscriptstyle a}; K_{\scriptscriptstyle a})$, where $\Pi(\cdot;\cdot)$ is an orthographic projection, we can re-project 3D keypoints to 2D keypoints.

Then, 
given the ground truth 2D keypoints, ${\mathbf{x}}_{\scriptscriptstyle {\{h,a\}}}$,
%
the full-keypoint reconstruction loss can be computed as the Euclidean distance between ${\mathbf{x}}_{\scriptscriptstyle{\{h,a\}}}$ and $\hat{\mathbf{x}}_{\scriptscriptstyle{\{h,a\}}}$: \vspace{-1mm}
\begin{equation}
    \small{L_{{\scriptscriptstyle full}}^{h} = \tfrac{1}{N_{\scriptscriptstyle h}}\textstyle{\sum_i v_{\scriptscriptstyle{h_i}} \lVert {\mathbf{x}}_{\scriptscriptstyle{h_i}} {-} {\hat{\mathbf{x}}_{\scriptscriptstyle{h_i}}}\rVert^{\scriptscriptstyle 2}_{\scriptscriptstyle 2}}, \quad L_{\scriptscriptstyle{full}}^{a} = \tfrac{1}{N_{\scriptscriptstyle a}}\textstyle{\sum_i v_{\scriptscriptstyle{a_i}}\lVert {\mathbf{x}}_{\scriptscriptstyle{a_i}} {-} {\hat{\mathbf{x}}_{\scriptscriptstyle{a_i}}}\rVert^{\scriptscriptstyle 2}_{\scriptscriptstyle 2}}},
    \vspace{-1mm}
\label{eq:fullkpt}
\end{equation}
where $N_{\scriptscriptstyle{\{h,a\}}}$ are the respective numbers of visible keypoints of humans and animals, ${\mathbf x}_{\scriptscriptstyle{\{h_i,a_i\}}}$ are the $i$-th human and animal ground truth 2D keypoints, respectively, and $v_{\scriptscriptstyle{\{h_i,a_i\}}}$
are the visibility for each 
ground truth keypoint (1 if visible, 0 otherwise). 
%
Besides,
we 
use the silhouette loss for animal classes, since animals show distinctive body shape difference across the classes. 
%
Given the animal binary mask annotation $S_{\scriptscriptstyle{a}}$ and the estimated 
binary mask $\hat{S}_{\scriptscriptstyle{a}}$, the silhouette loss can be computed by
the negative IoU~\cite{kato2018renderer,liu2019soft} as 
${L_{\scriptscriptstyle sil} = 1 {-} \tfrac{\lVert \hat{S}_{\scriptscriptstyle{a}}{\scriptscriptstyle \otimes} S_{\scriptscriptstyle{a}}\rVert_{\scriptscriptstyle 1}}{\lVert \hat{S}_{\scriptscriptstyle{a}}{\scriptscriptstyle \oplus} S_a-\hat{S}_a{\scriptscriptstyle \otimes} S_a\rVert_1}}$,
where $\oplus$ and $\otimes$ denote the respective element-wise sum and product.

Given an image of a class, \eg, human, the described 
modules and losses
are only partially available due to the absence of annotations of another branch, \eg, animals; thus, the other branch is not jointly trainable.
Hence, we introduce our novel components of \DeMR\ that relax original disjoint multi-task problem into a jointly trainable scheme.

\vspace{-2mm}
\subsection{Morphological Similarity based Multi-task Learning} \label{sec:morph}
\vspace{-1mm}
Our \DeMR\ becomes jointly trainable
under the following hypothesis: the \emph{morphological similarity} across different classes of deformable objects provides additional knowledge and it can mitigate the difficult optimization in training.
%
We first define the morphological similarity such that the effectiveness of hypothesis is realized,
and then present the techniques that can induce more distinctive outputs specific for each class branch.
%

\begin{wrapfigure}{R}{0.53\textwidth}
    \centering
    \vspace{-5mm}
    \includegraphics[width=\linewidth]{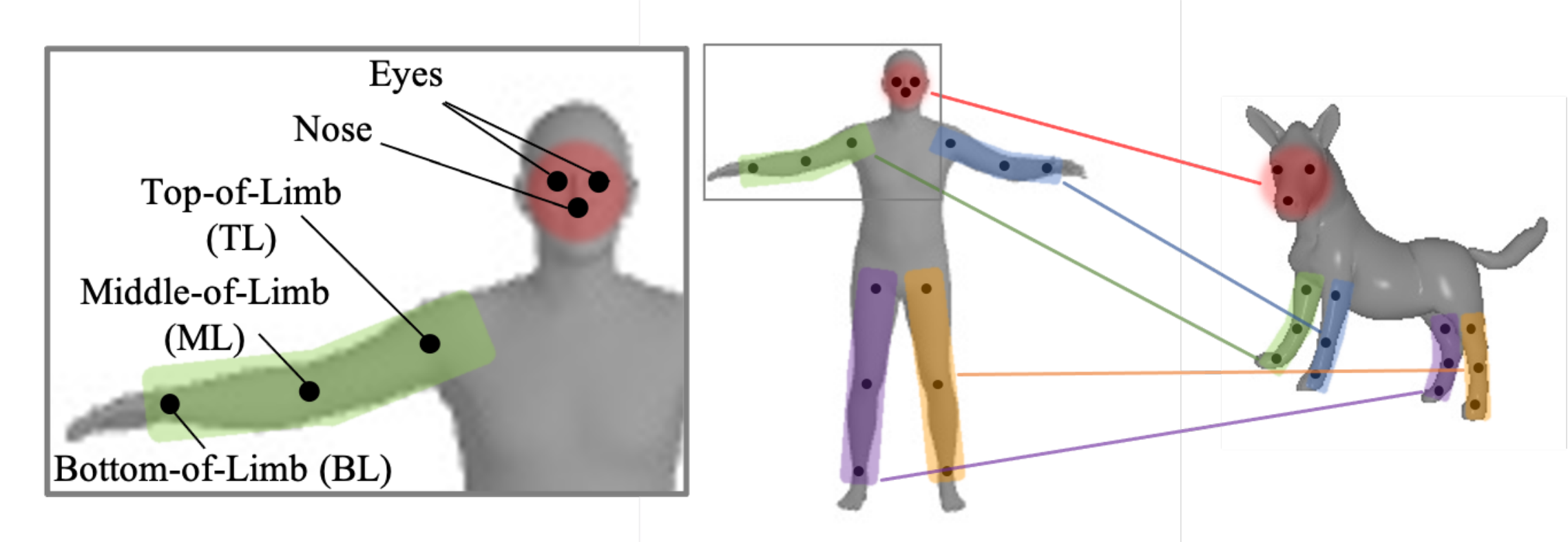}
    \vspace{-3mm}
    \caption{{\textbf{Sub-keypoint groups and pairs.} The same colored regions correspond to morphologically similar ones across heterogeneous classes.}}
  \vspace{-5mm}
\label{fig:subkpt}
\end{wrapfigure}

\para{Morphological Similarity} 
Our observation is that physical corresponding body parts exist among different species, such as arms and legs.
%
We use the keypoint of \emph{Eyes, Nose, Top-of-Limb (TL), Middle-of-Limb (ML), Bottom-of-Limb (BL)} on the arms, the legs, and the head, called sub-keypoints (see \Fref{fig:subkpt}). 
We force 
the model to reconstruct an object class with a body model of a different class, \ie, mimicking a pose like the animal exercise.
We induce it by minimizing the difference between 
the supervision of sub-keypoints of a class and the estimated sub-keypoint of the other class.
Thereby, the model learns morphological similarity among heterogeneous classes. 
We empirically select $N_{\scriptscriptstyle sub}$ keypoints from the ground truth full 2D keypoints and define them as sub-keypoints, \ie, $\mathbf{x}_{\scriptscriptstyle \{h, a\}}^{\scriptscriptstyle{sub}}\subset{\mathbf{x}_{\scriptscriptstyle \{h, a\}}}$ 
(refer to the supplementary materials for details).

\para{Sub-keypoint based Multi-task Learning}
Introducing the sub-keypoint concept allows 
joint 
learning of multi-task branches in the incompatible labeled set setup.
%
%
Specifically, given the example with a horse in an input image
in \Fref{fig:network}, the model tries to estimate both SMPL and SMAL parameters in each branch.
However, since human keypoint annotations do not exist for a horse image,
the model cannot compute the full keypoint loss $\textstyle{L_{\scriptscriptstyle full}^{h}}$,
and discards the predictions from the human 
branch. 
This wastes that branch, and 
leads to the forgetting phenomenon during training~\cite{li2017learning}, resulting in unstable
multi-task learning.
%
%
Our sub-keypoint reconstruction loss for inducing morphological similarity can resolve this limitation of the na\"ive multi-task learning.
Since the loss enforces the estimated sub-keypoints of the human SMPL mesh to follow the morphologically corresponding sub-keypoint ground truth of a horse; thus, even without human labels for the horse image, we can compute the loss for the human mesh branch as well and get gradients during the backpropagation, enabling stable joint training of both branches.
This is \emph{vice versa} for a human image.
 
The computation of the sub-keypoint reconstruction loss is identical to that of the full-keypoint loss, except it uses the $N_{\scriptscriptstyle sub}$ morphologically corresponding sub-keypoints (we set $N_{\scriptscriptstyle sub} {=} 10$).
%
Given the ground truth sub-keypoints
$\mathbf{x}_{\scriptscriptstyle \{h, a\}}^{\scriptscriptstyle{sub}}\subset{\mathbf{x}_{\scriptscriptstyle \{h, a\}}}$,
estimated sub-keypoints 
${\hat{\mathbf{x}}}_{\scriptscriptstyle \{h, a\}}^{\scriptscriptstyle{sub}}\subset{\hat{\mathbf{x}}}_{\scriptscriptstyle \{h, a\}}$ and the visibility ${v}_{\scriptscriptstyle \{h, a\}}^{\scriptscriptstyle{sub}}$,
the sub-keypoint 
loss
can be computed 
as follows:\vspace{-1mm}
\begin{equation}
    \small{L_{\scriptscriptstyle sub}^{h} {=} \tfrac{1}{N_{\scriptscriptstyle sub}}\textstyle{\sum_{i=0}^{N_{\scriptscriptstyle sub}} \lVert {v}_{\scriptscriptstyle h_i}^{\scriptscriptstyle{sub}}({\mathbf{x}}_{\scriptscriptstyle h_i}^{\scriptscriptstyle{sub}} {-} {\hat{\mathbf{x}}_{\scriptscriptstyle a_i}^{\scriptscriptstyle{sub}}})\rVert^{\scriptscriptstyle 2}_{\scriptscriptstyle 2}}, \quad
    L_{\scriptscriptstyle sub}^{a} {=} \tfrac{1}{N_{\scriptscriptstyle sub}}\textstyle{\sum_{i=0}^{N_{\scriptscriptstyle sub}} \lVert {v}_{\scriptscriptstyle a_i}^{\scriptscriptstyle{sub}}({\mathbf{x}}_{\scriptscriptstyle a_i}^{\scriptscriptstyle{sub}} {-} {\hat{\mathbf{x}}_{\scriptscriptstyle h_i}^{\scriptscriptstyle{sub}}})\rVert^{\scriptscriptstyle 2}_{\scriptscriptstyle 2}}}.
\label{eq:subkpt}
\end{equation}
The effect of sub-keypoint loss is shown in mesh estimation results in \Fref{fig:network}. Since the model is supervised with the ground truth sub-keypoint of a horse, it eventually estimates a human SMPL mesh that mimics the pose and shape of the horse.


\begin{wrapfigure}{r}{0.42\linewidth}
\centering
    \vspace{-4mm}
    \includegraphics[width=\linewidth]{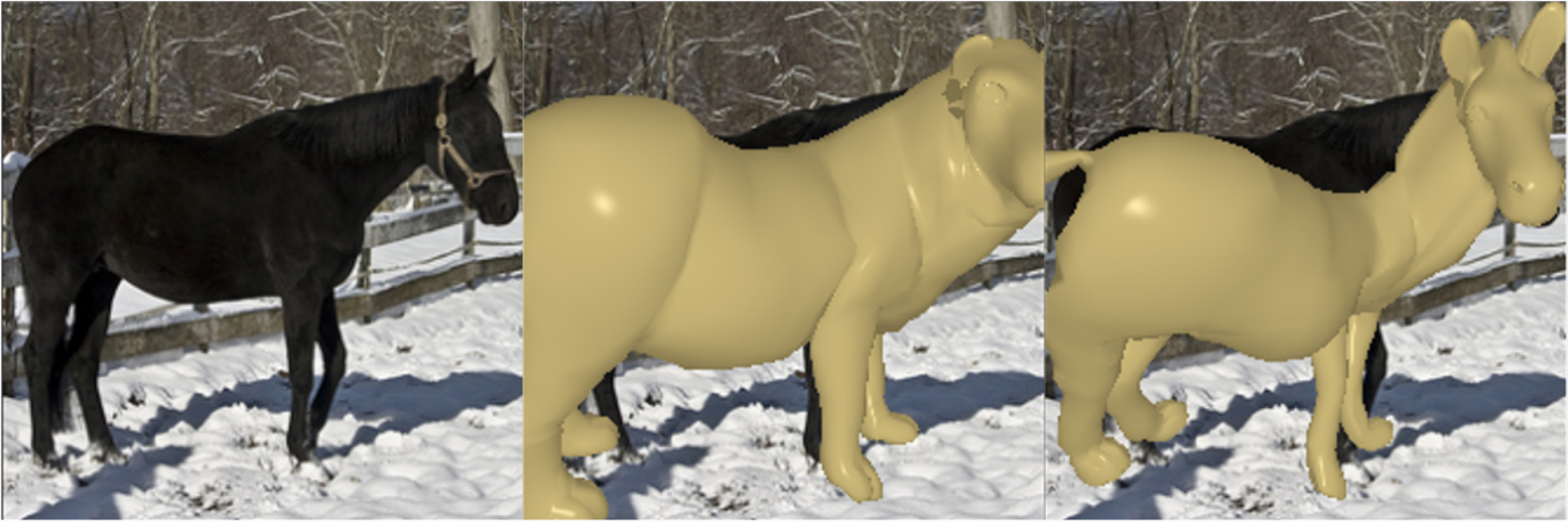}
    \vspace{-2mm}
    \caption{{\textbf{Effect of the class-selective shape prior.} 
    The models trained with uni-modal~(middle) and class-selective shape prior~(right).}
    }
    \vspace{-4mm}
\label{fig:shapeprior}
\end{wrapfigure}

\para{Class-selective Shape Prior Loss}
As shown in the prior works~\cite{biggs2020wldo,kolotouros2019spin},
regularizing the estimated shape parameters to follow 
the natural distribution of each class is essential in achieving realistic bodies of deformable objects. 
%
Since different animal classes show distinctive shape parameter distributions, we extend the uni-modal shape prior to
the \emph{class-selective shape prior loss} to adaptively regularize shape parameters according to
the target animal class.
%
Given a multivariate Gaussian shape prior distribution of the animal class, $c$, with a mean $\textstyle{\mu}_{\scriptscriptstyle{\beta_a}}^{c}$ and a covariance 
$\textstyle{\Sigma}_{\scriptscriptstyle{\beta_a}}^{c}$, the 
shape prior loss can be computed by the
Mahalanobis distance for 
the predicted animal shape parameters, ${\textstyle{\beta_a}}$, as: 
\vspace{-1mm}
\begin{equation}
    \small{L_{\scriptscriptstyle sh}^{a}{=}
    \textstyle{\sum_{c=0}^{N_{\scriptscriptstyle animals}}} \mathbbm{1}[y=c] \cdot (\beta_a{-}\mu_{\scriptscriptstyle{\beta_a}}^{c})^\top\inv{( \Sigma_{\scriptscriptstyle{\beta_a}}^{c})}(\beta_a{-}\mu_{\scriptscriptstyle{\beta_a}}^{c})},
    \vspace{-1mm}
\end{equation}
where $y$ denotes the ground truth animal class label, $\mathbbm{1}[\cdot]$ the indicator function that returns 1 for true and 0 otherwise.
Since we aim
to estimate realistic 3D mesh of three different animal classes which have class-wise different shape prior means and covariance matrices, the proposed 
class-selective shape prior loss $L_{\scriptscriptstyle sh}^{a}$
can reflect class-specific statistics of shape given the ground truth animal class label $y$ in training time.
%
The effect of class-selective 
shape prior loss is shown
in \Fref{fig:shapeprior}, where the model trained with the uni-modal prior estimates mean-shaped 
animal mesh,
while the model with our prior estimates a more horse-like shape.

\para{Class-specific Batch Normalization (CSBN)}
As mentioned, there are statistical gaps between humans and animals in shared feature vectors obtained from the feature extractor.
This is detrimental for multi-task branches because the mismatch is likely to be propagated to the mesh parameter output layer.
To take into account the statistical gap, we adopt two different batch normalization layers in the front of each branch, denoted CSBN.
CSBN enables the model to learn class-specific residual information and statistical gaps among heterogeneous class data
while allowing the model to share all other model parameters, \ie, CNN backbone. 
There is a similar work 
called domain-specific batch normalization~\cite{chang2019domainspecific}, but different from ours in that it switches the statistics at a single path according to an input domain label but ours is deployed for adapting statistics for independent branching.



\section{Experiments}
\vspace{-2mm}
In this section, we evaluate our proposed
\DeMR\, in several aspects. 
Specifically, we first introduce datasets and evaluation metrics 
for training and evaluation in \Sref{Sec:datasets}. 
We then evaluate our method by comparing it with the previous uni-modal 
methods
in \Sref{Sec:evaluation}.
%
Lastly, we further analyze 
the effects of the proposed key components
in \Sref{Sec:ablation}.

\vspace{-2mm}
\subsection{Experiment Setup}   \label{Sec:datasets}
\vspace{-1.5mm}
Unlike human datasets that have diverse annotations, including 2D, 3D keypoints, or segmentation mask annotations,
supervised datasets themselves are rare for animals; 
moreover, real 3D keypoint annotations are extremely difficult to obtain and far rarer for animals. 
Considering practical usages, we focus on the weakly-supervised case with 2D keypoint datasets for both humans and animals.
%
%
Concretely, we follow the evaluation setup same with ~\cite{kolotouros2019spin, biggs2020wldo}. We use LSP~\cite{Johnson10}, LSP-extended~\cite{Johnson11}, MS COCO~\cite{lin2015microsoft}, and MPII~\cite{andriluka14cvpr} for training the human domain, and use Stanford Extra~\cite{biggs2020wldo} and Animal Pose~\cite{Cao2019CrossDomainAF} for training the animal domain.
For evaluation, we employ Human3.6M~\cite{h36m_pami}, MPI-INF-3DHP~\cite{mono-3dhp2017}, 3DPW~\cite{vonMarcard2018} as the human domain evaluation sets, and Stanford Extra and Animal Pose for the animal domain evaluation sets. 
We report the Procrustes-aligned mean per joint position error (PA-MPJPE), the mean per joint position error~(MPJPE) and the Percentage of Correct Keypoints~(PCK). 

Note that even though the model is trained with 2D keypoint datasets in both human and animal domains, we evaluate \DeMR's 3D recovery performance in the human domain to report our method's favorable 
performance with the competing methods.

%
\vspace{-2.5mm}
\subsection{Evaluation}
\label{Sec:evaluation}
\vspace{-1.5mm}
\para{Competing Methods}
Since our model is the first approach that addresses the 3D mesh reconstruction of both humans and animals using a unified model, no prior works are directly comparable.
%
We first compare with the base neural models that we mainly refer to and are the closest comparators, \ie, HMR~\cite{hmrKanazawa17} for humans and WLDO~\cite{biggs2020wldo} for dog breeds. 
Then, we compare our unified method with more recent uni-modal models for humans.
%
%
%
%

%

\begin{table*}[t]
    \centering
    \resizebox{\columnwidth}{!}{%
    \begin{tabular}{l c c c c c c c}
        \toprule
        & \multicolumn{2}{c}{} & \multicolumn{1}{c}{MPJPE [mm] $\downarrow$} & \multicolumn{2}{c}{PA-MPJPE [mm] $\downarrow$} & \multicolumn{2}{c}{PCK [\%] $\uparrow$} \\\cmidrule(lr){2-3} \cmidrule(lr){4-4} \cmidrule(lr){5-6} \cmidrule(lr){7-8}
        \multirow{2}{*}{Models}& \multirow{2}{*}{\# of Params.} & \multirow{2}{*}{Trained Object Classes} &  MPI-INF-3DHP & Human3.6M-P2 & 3DPW & Stanford Extra & Animal Pose \\ 
        
        & &  & (Human) & (Human) & (Human) & (Dog) & (Horse, Cow) \\

        \cmidrule{1-8}
        
        HMR~\cite{hmrKanazawa17} & 27M & Human & 169.50 & 66.50 & 81.30 & - & -\\
        WLDO~\cite{biggs2020wldo} & 95M & Dog & - & - & - & 78.80 & 41.05 \\
        
        \cmidrule{1-8}
        
        Ours (4cls, na\"ive) & \multirow{3}{*}{27M+5M} & \multirow{3}{*}{Human, Dog, Horse, Cow} & 154.57 & 81.01 & 70.74 & 72.79 & 51.14 \\
        \quad \emph{+ Sub} & & & 155.34 & 82.42 & 69.94 & 72.90 & \underline{\textbf{52.72}} \\
        \quad \emph{+ Sub + CSBN (full)} &  & &\underline{\textbf{140.76}} & \underline{\textbf{79.70}} & \underline{\textbf{69.85}} & \underline{\textbf{73.23}} & 50.09 \\
        
        \cmidrule{1-8}
        
        Ours (2cls, na\"ive) & \multirow{2}{*}{27M+5M} & \multirow{2}{*}{Human, Dog} & 160.14 & 81.13 & \underline{\textbf{68.75}} & 73.87 & - \\
        \quad \emph{+ Sub + CSBN} & &  & \underline{\textbf{156.69}} & \underline{\textbf{80.28}} & 70.64 & \underline{\textbf{74.29}} & - \\
        
        \bottomrule
    \end{tabular}
    }
    \vspace{2mm}
    \caption{Quantitative evaluation on various human/animal datasets.}
    \label{table:all}
    \vspace{-1mm}
\end{table*}

\begin{table}[t]
    \centering
    \resizebox{\columnwidth}{!}{
        \begin{tabular}{c@{}c}
            \begin{minipage}{0.8\columnwidth}
                \centering
                \begin{tabular}{l c}
                    \toprule
                    \multicolumn{1}{c}{} & \multicolumn{1}{c}{PA-MPJPE [mm] $\downarrow$}
                    \\ \cmidrule(lr){1-2}
                    HMR~\cite{hmrKanazawa17} & 81.3  \\
                    Kanazawa \etal~\cite{humanMotionKanazawa19}  & 72.6   \\
                    Kolotouros \etal~\cite{kolotouros2019cmr} & 70.2
                    \\ \cmidrule(lr){1-2}
                    Ours       & \underline{\textbf{69.85}} \\ 
                    \bottomrule
                \end{tabular}\\
                \vspace{1mm}
            \end{minipage} &
        
            \begin{minipage}{0.8\columnwidth}
                \centering
                \begin{tabular}{l c}
                    \toprule
                    \multicolumn{1}{c}{} & \multicolumn{1}{c}{PA-MPJPE [mm] $\downarrow$}
                    \\ \cmidrule(lr){1-2}
                    Lassner \etal~\cite{Lassner:UP:2017}  & 93.9   \\
                    Pavlakos \etal~\cite{Pavlakos2018LearningTE}  & 75.9\\
                    HMR~\cite{hmrKanazawa17} & 66.5
                    \\ \cmidrule(lr){1-2}
                    Ours       & 79.7 \\ 
                    \bottomrule
                \end{tabular}%
                \vspace{1mm}
            \end{minipage} \\
        \small (a) & \small (b)
        \end{tabular}
    }
\vspace{1.5mm}
\caption{(a) Evaluation on 3DPW dataset. (b) Evaluation on Human3.6M dataset P2.}
\vspace{-3mm}
\label{table:human}
\end{table}

\para{Quantitative Evaluation}
Tables~\ref{table:all} and~\ref{table:human} provide detailed comparisons 
to other methods 
on several
3D human datasets. 
While our base structure is the same with HMR~\cite{hmrKanazawa17}, \DeMR\ show improved performances on MPI-INF-3DHP and 3DPW, but has 
lower performance on Human3.6M. 
We postulate 
that a huge domain gap between our training datasets of humans and animals and Human3.6M led to 
performance degradation. 
Remind that \DeMR\ is trained with 2D outdoor human and animal datasets~\cite{andriluka14cvpr,Johnson10,Johnson11,lin2015microsoft}. 
The MPI-INF-3DHP and 3DPW datasets, where our method performs better, mainly consist of outdoor human motion sequences similar to our training datasets.
On the other hand, Human3.6M 
is captured in controlled indoor scenes, which shows totally different characteristics, including backgrounds, 
human poses, 
and light conditions.
Particularly, \DeMR\ leverages animal exercise, whereby our method may be more favorable for unusual poses rather than indoor poses.  
%

We further compare with other competing methods in \Tref{table:human}, where our model favorably performs against the uni-modal approaches specifically designed for humans.
%
Note that our model performs 
better 
than Kanazawa~\etal~\cite{humanMotionKanazawa19}, which is a video-based multi-frame approach to deal with temporally consistent human motions, \ie, unfair to ours.
The result supports our hypothesis
that the morphological similarity among humans and animals 
gains to obtain complementary information and 
powerful reconstruction in a single unified model.


\begin{figure*}[t]
    \centering
        \includegraphics[width=1.0\linewidth]{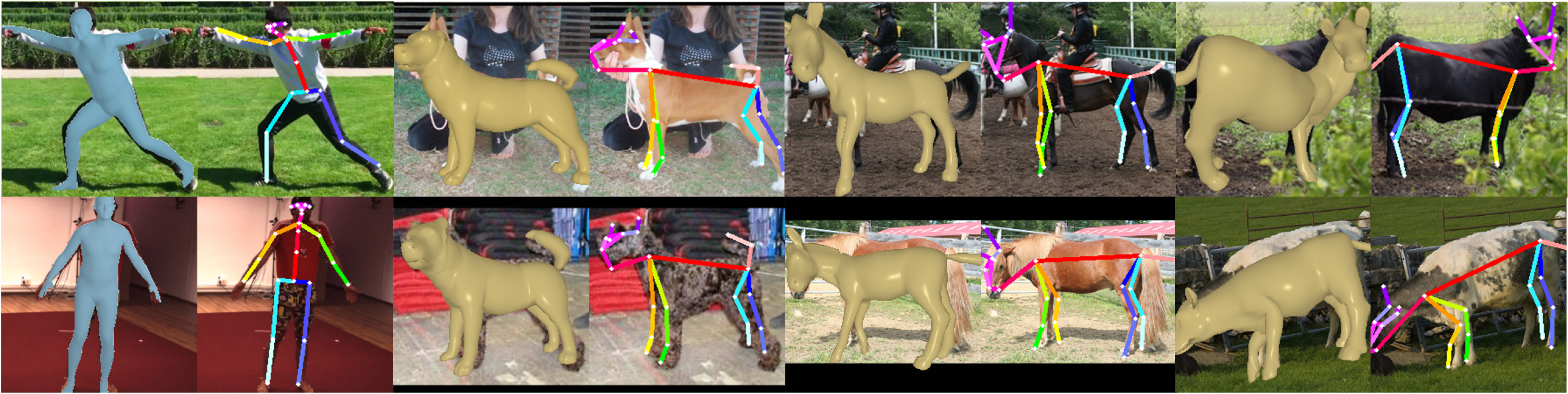}\\
        \vspace{3mm}
        \caption{{\textbf{Qualitative results on multiple datasets.} Each column shows the result of predicted mesh and predicted keypoints alternatively for human, dog, horse, and cow classes. The images are from 3DPW, Human3.6M (human), Stanford Extra (dog) and Animal Pose dataset (horse, and cow).}}
        \vspace{-6mm}
\label{fig:allresult}
\end{figure*}

Regarding the animal classes, 
%
WLDO is the strong competing method because it is designed and tuned for dog breeds and is evaluated on Stanford Extra that is a dog-only dataset.
In \Tref{table:all}, our \DeMR\ shows slightly lower performance than that of WLDO, but
%
this is still favorable performance in that, from HMR, our method adds up a new capability to deal with more diverse animals with a small increase of the model size over that of HMR.
\DeMR\ is three times smaller than the model size of WLDO, and can deal with three additional animals, \ie, humans, horses, and cows. 
%
This result implies that the morphological
similarity among different classes helps our model cover more classes while preserving 
the uni-modal model's performance with much compact networks. 

For Animal Pose,
we use
the horse and cow categories as the test set. 
Although WLDO also evaluates the model on Animal Pose, they only use the dog category as a test set since WLDO only covers dogs. 
Note that there is no prior work that directly measures PCK on diverse animal classes as ours.
Therefore, we mainly compare our full model with our simpler versions as an ablation study,
\ie,~ours with/without CSBN 
and with/without the sub-keypoint loss, which will be discussed in \Sref{Sec:ablation}. We additionally report the WLDO's performance measured on other Animal Pose classes. Since WLDO has the most similar structure with ours that reconstructs 3D animal meshes, and can be tested on other quadruped animals, we additionally test WLDO on other proximal animal classes for broader evaluation. It is not surprising that \DeMR\ shows about 10\% higher PCK than WLDO, since WLDO did not see horse and cow images in training time. The result shows \DeMR's ability to cover more animal classes comparably even with smaller network capacity.
Also, by the fact that our two-class model has better performance than our four-class one, it shows that there exists a trade-off relationship according to the number of animal classes to be dealt with.


\para{Compactness}
As shown in \Tref{table:all}, the main advantage of \DeMR\ is 
its competitive performance to uni-modal models while having much compact architecture in terms of the scale of network parameters. We highlight that our model has a smaller number of network parameters (27M+5M) compared to the sum of independent networks 
(27M+95M) of HMR and WLDO while covering 5 times broader classes of objects than HMR and WLDO.

\begin{figure*}[t]
    \centering
        \includegraphics[width=1.0\linewidth]{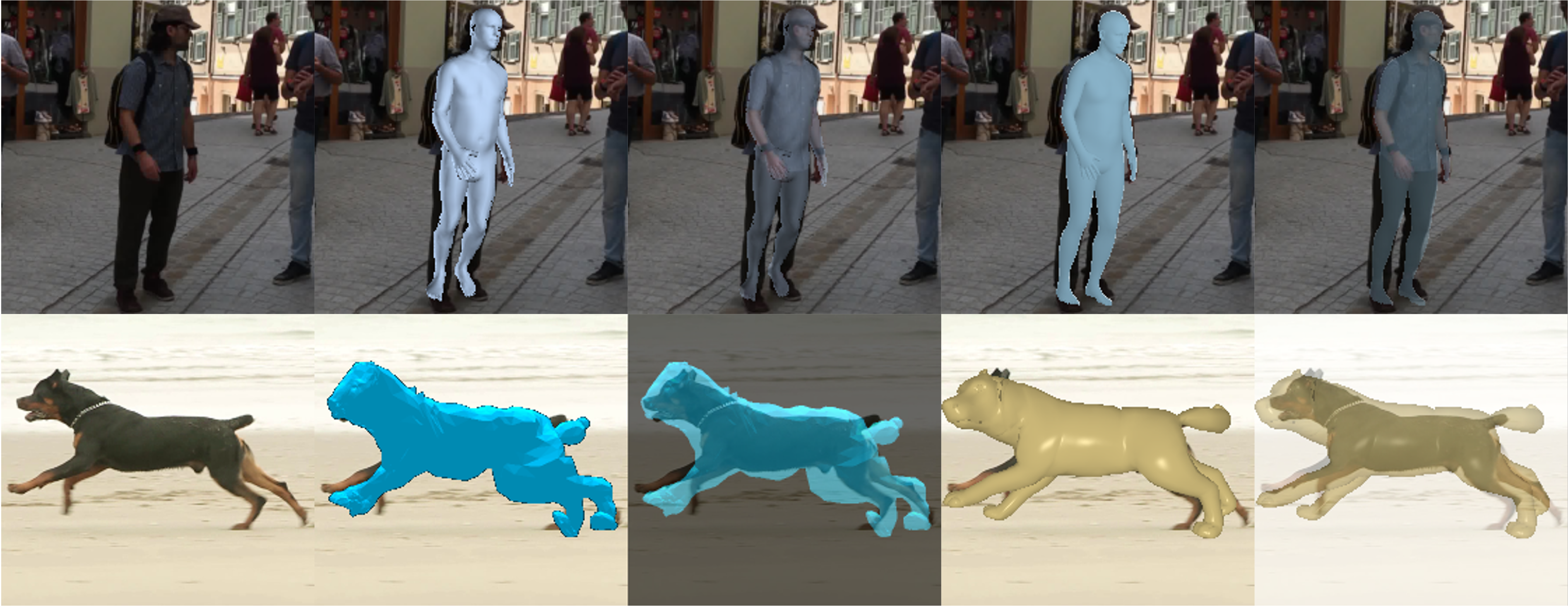}\\
        \footnotesize{(a) \hspace{2.1cm} (b) \hspace{2.1cm} (c) \hspace{2.1cm} (d) \hspace{2.1cm} (e)}
        \vspace{3mm}
        \caption{\textbf{Qualitative comparison with competing methods.} 
        (a) input image, (b) mesh prediction and (c) mesh silhouette of competing methods (top: HMR~\cite{hmrKanazawa17}, bottom: WLDO~\cite{biggs2020wldo}), (d) \DeMR's mesh prediction and (e) \DeMR's mesh silhouette for benchmarks, respectively.
        }
        \vspace{-6mm}
\label{fig:base_comp}
\end{figure*}

\para{Qualitative Results}
In \Fref{fig:allresult}, we visualize \DeMR's 3D mesh estimation for 
humans and animals
and estimated joint keypoints regressed from the estimated meshes. 
It shows plausible results for every target class and distinctive shapes 
for different species. Figure \ref{fig:base_comp} shows the comparison results of \DeMR\ to its competing methods, HMR and WLDO. Qualitative improvements can be found in parts, such as the limbs of the human and the dog. More qualitative results can be found in supplementary materials.
%

\vspace{-4mm}

\subsection{Ablation study} \label{Sec:ablation}
\vspace{-1mm}
\begin{wrapfigure}{r}{0.40\textwidth}
\centering
    \vspace{-4mm}
    \includegraphics[width=\linewidth]{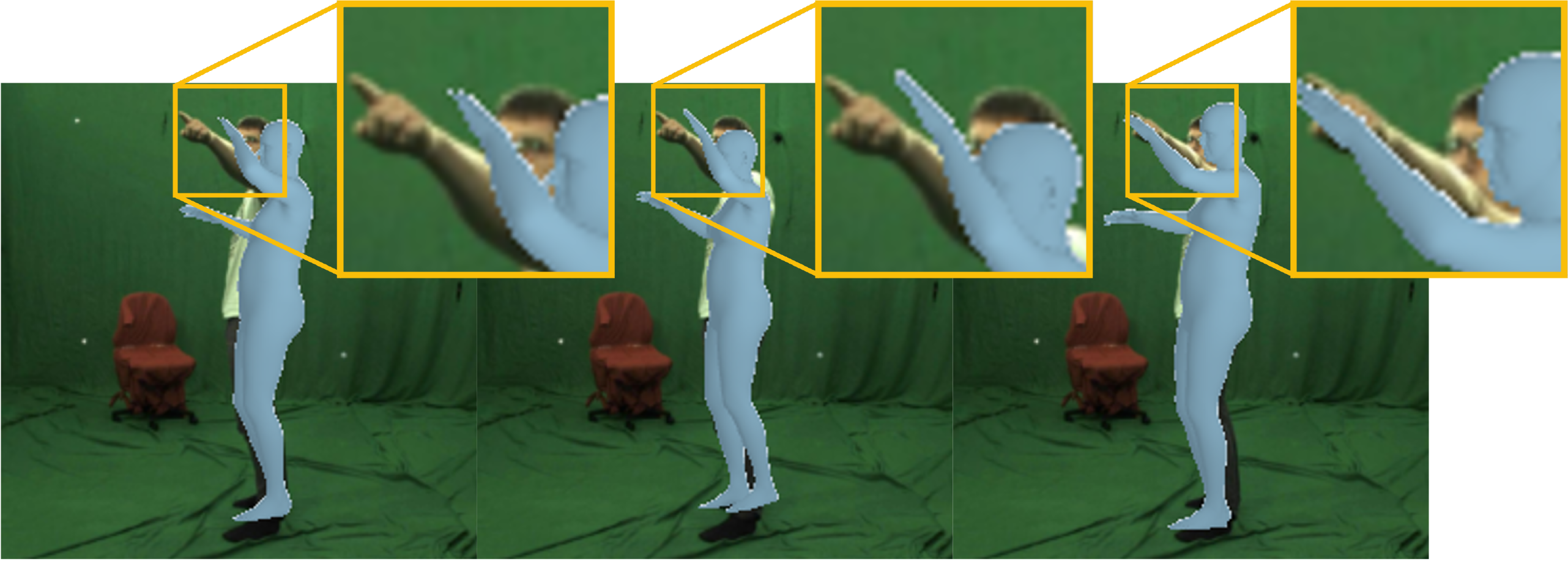}
\vspace{0.25mm}
\caption{\textbf{Qualitative ablation.} Predicted mesh results of the models. Na\"ive (left),  na\"ive with sub-keypoint (mid), and full (right).}
\vspace{-4mm}
    \label{fig:ablation_fig}
\end{wrapfigure}

We ablate \DeMR\
with different settings of the loss and the architecture. We define our full model as the one that uses CSBN and 10 sub-keypoint pairs in loss computation.
Also, the na\"ive multi-task model is the one that uses no CSBN and no
sub-keypoint pairs.
Figure \ref{fig:ablation_fig} shows the predicted mesh comparison across ablated models. In detail, the mesh predicted by the na\"ive model shows the poorest quality, while adding sub-keypoint and CSBN in training significantly improves the reconstruction quality.

\begin{wraptable}{R}{0.58\textwidth} 
\vspace{-5mm}
    \centering
        \begin{tabular}{c@{}c} 
        \resizebox{0.69\linewidth}{!}{%
        \begin{tabular}{cc ccccc cc} 
        \toprule
             & & \multicolumn{5}{ c }{{\bfseries Labels of Sub-Keypoints}} &\multicolumn{2}{c}{Class: Dog}\\ 
            \cmidrule(lr){3-7}
            \multicolumn{2}{c}{\# of Subkpt}
            & Eyes & BL & TL & ML & Nose
            & \multicolumn{2}{c}{PCK [\%] $\uparrow$}\\ 
            \midrule
            \multicolumn{2}{c}{15} & \ccy{\checkmark} & \ccy{\checkmark} & \cco{\checkmark} & \ccr{\checkmark} & \ccr{\checkmark} & \multicolumn{2}{c}{73.16}\\
            \multicolumn{2}{c}{\textbf{\underline{10}}} & \ccy{\checkmark} & \ccy{\checkmark} & \cco{\checkmark} & & & \multicolumn{2}{c}{\textbf{\underline{73.23}}}\\ 
            \multicolumn{2}{c}{6}  & \ccy{\checkmark} & \ccy{\checkmark} & & & & \multicolumn{2}{c}{72.14}\\ 
            \multicolumn{2}{c}{0}  & & & & & & \multicolumn{2}{c}{71.89}\\ 
        \bottomrule
        \end{tabular}}&\hspace{-2mm}
            \raisebox{-.45\height}{\includegraphics[width=0.29\linewidth]{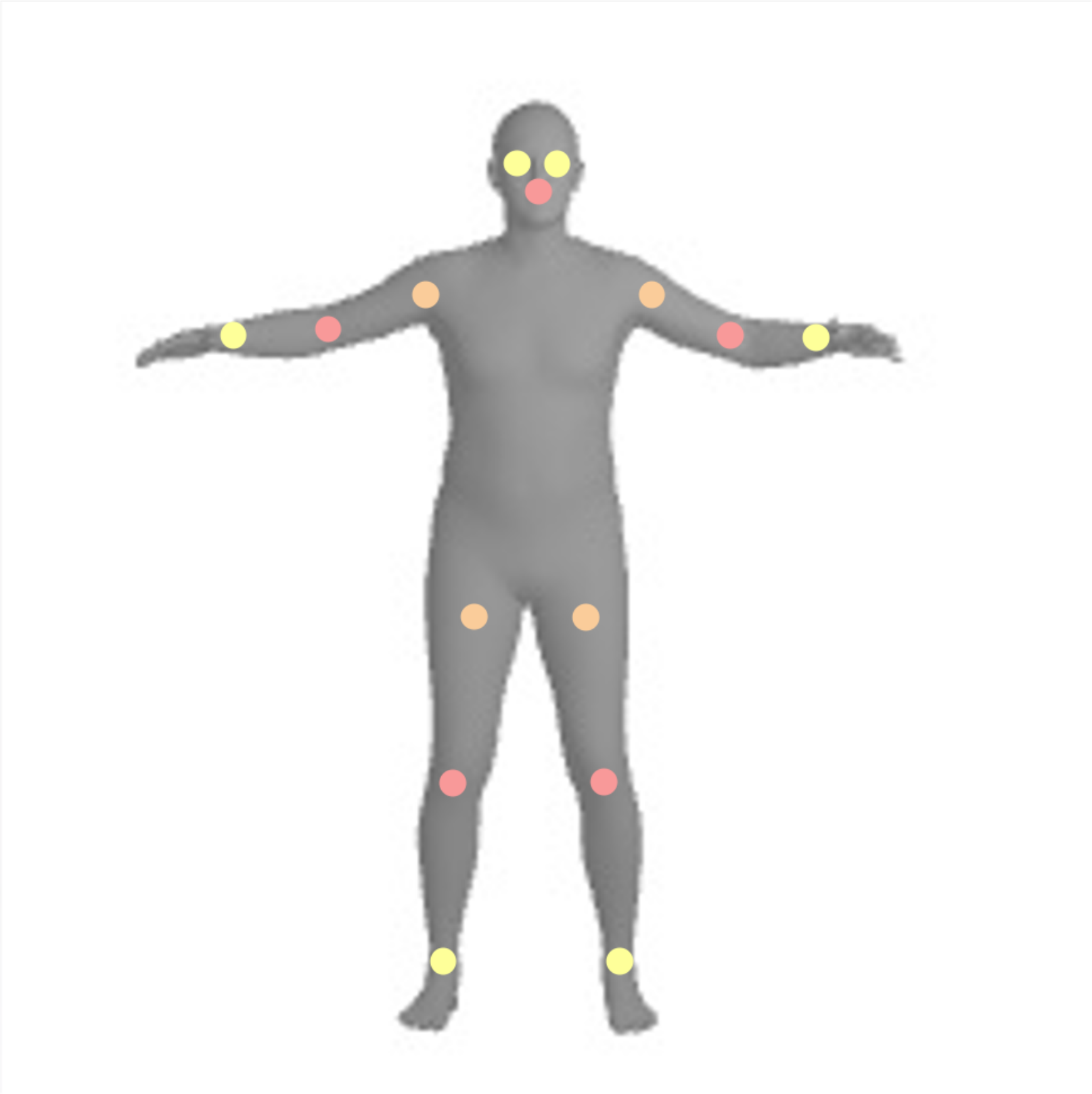}} \hspace{-2mm}
            \\
        \small (a) & \small (b)
        \end{tabular}
        \vspace{2mm}
        \caption{\textbf{Quantitative ablation.} (a) PCK measured on various sub-keypoint numbers. (b) Sub-keypoint groups denoted by colors.}
        \label{table:ablation:subkpt}
        \vspace{-5mm}
\end{wraptable}
\vspace{2mm}
\para{Effects of $L_{\scriptscriptstyle sub}^{*}$}
We evaluate how many sub-keypoints we need to enable
the model to learn the morphological similarity. 
\Tref{table:ablation:subkpt} shows PCK evaluated
on Stanford Extra~\cite{biggs2020wldo} according to the 
different numbers 
of sub-keypoints
used in {\small $L_{\scriptscriptstyle sub}^{*}$}.
We found that the model using 10 sub-keypoints achieves the best performance on animal reconstruction. Using 15 sub-keypoint pairs in training
turns out disturbing, 
while 
using 6 sub-keypoint pairs lacks the morphological information to be learned properly. 
The model that uses no sub-keypoint in training
shows the lowest performance. 
Simply adding sub-keypoint loss on the na\"ive multi-task model improves the reconstruction performance on heterogeneous classes.
More analysis and discussion can be found in the supplementary material.


\para{Effects of CSBN}
In \Tref{table:all}, we compare our full model with the model without CSBN.
Adding CSBN 
enhances the reconstruction performance on almost every dataset except Animal Pose.
All the other datasets cover
one target class, 
while Animal Pose 
consists of multiple animal classes. 
We analyze this result as the limitation stemming from
the simple deployment of CSBN only in front of the human and animal branches.
Thus, CSBN deals with 
the inter-class statistical difference only,
\ie, humans and animals, and restricts the statistics to be similar across quadruped animals by a single CSBN.
We postulate that a fine-grained extension of CSBN like DSBN at the animal branch may further improve the performance for multiple animals.



\begin{wraptable}{R}{0.45\textwidth}    
    \vspace{-2mm}
    \centering
    \resizebox{0.45\columnwidth}{!}{%
        \begin{tabular}{cc ccccc c} 
        \toprule
            & & \multicolumn{4}{ c }{{\bfseries Number of datasets available}} &\multicolumn{2}{c}{Class: Dog}\\ 
            \cmidrule(l){3-6}
          \multicolumn{2}{c}{Dataset \#}& Human & Dog & Horse & Cow & PCK [\%] $\uparrow$\\ 
            \midrule
            \multicolumn{2}{c}{0.6K} & 0.3K & 0.3K & - & - & 40.83\\ 
            \multicolumn{2}{c}{\textbf{\underline{1.2K}}} & 0.3K & 0.3K & 0.3K & 0.3K & \textbf{\underline{48.07}}\\ 
            \multicolumn{2}{c}{1.0K} & 0.3K & 0.1K & 0.3K & 0.3K & 38.50\\ 
            \multicolumn{2}{c}{\textbf{\underline{1.2K}}} & 0.5K & 0.1K & 0.3K & 0.3K & \textbf{\underline{43.64}}\\ 
        \bottomrule
        \end{tabular}%
        }
        \vspace{2mm}
        \caption{\textbf{Quantitative results.} Sample efficiency experiments on Stanford Extra.}
        \label{table:ablation:dataset}
        \vspace{-4mm}
\end{wraptable}


\para{Sample Efficiency} 
We evaluate the sample efficiency of \DeMR\, in
extremely low training data scenarios.
Suppose 
one has a limited amount of dog data, \eg, about 300 images, and wants to reconstruct 3D dogs reasonably.
By virtue of our disjoint multi-task learning of \DeMR\, 
we can
leverage 
other species' data, including humans and quadruped animals.
%
\Tref{table:ablation:dataset} shows the results on several low-data regimes.
While the model trained with dog data only has poor performance, 
the model trained with both human and dog datasets starts to have plausible PCK  
of 
40.83\%. 
To fully exploit the 
morphological similarity, simply adding other quadruped animal data in 
training improves
the performance by 
7.2\%.

In the case of a smaller number of training data, \eg, 100 dog images,
PCK significantly decreases by 
9.57\% even with other class data. 
While not being as much power as adding additional data of the same quadruped animals, adding 200 more non-quadruped animal data, \ie, human data, 
in 
training improves 
the performance.
This shows
our method of learning the morphological similarity is effective 
even in low-data scenarios; 
thus, sample efficient.
\vspace{-5mm}


\vspace{2mm}
\section{Conclusion}
\vspace{-1.5mm}
We present \DeMR, a compact and unified 3D mesh recovery of deformable objects. To deal with heterogeneous classes, we identify the challenge of disjoint multi-task learning, and relax it to be jointly trainable by introducing morphological similarity. 
We embody it by proposing the concept of sub-keypoint and class-aware regularizations.
Our \DeMR\ encompasses broad classes with competitive performance compared to the uni-modal models despite its compactness and sample-efficiency of our model. 
\DeMR's next attention is on reconstructing broader animal classes with extreme morphological characteristics, such as elephants and giraffes. 
We believe that
adopting implicit body representation~\cite{saito2020pifuhd} or neural-parametric model-based body reconstruction~\cite{palafox2021npms} along with our proposed sub-keypoint would be 
promising future directions. Moreover, extending \DeMR\ to temporally coherent mesh regression~\cite{kocabas2019vibe} would be mandatory future research direction to enable comprehensive motion analysis of deformable objects.

\clearpage
\para{Acknowledgment}
This work was carried out as a part of the deformable object recognition technology research project supported by the Agency for Defense Development, Korea, and by the Defense Acquisition Program Administration, Korea (UD200025ID).
This work was partially supported by the NIA grant and the IITP grant funded by the Korea government~(MSIT) (Artificial Intelligence Graduate School Program; [No. 2019-0-01906, POSTECH], [No. 2020-0-01336, UNIST]).
The GPU resource is supported by a study on the ``HPC Support'' project, supported by the MSIT and NIPA.
%
\bibliography{egbib}
\end{document}


\maketitle

The goal of this supplementary material is to provide additional contents and details that are not included in the main paper due to 
space constraints.
In \Sref{sec:subkeypoint}, we provide additional 
analysis on our proposed \emph{sub-keypoint}.
In \Sref{sec:quan_eval}, we quantitatively evaluate \DeMR\ with increased model capacity, showing promising further performance improvement.
In \Sref{sec:qual_eval}, we qualitatively compare 
our method with the animal domain competing method, 
WLDO on dog.
In \Sref{sec:traning_details}, we describe our training details, including hyper-parameters and training hardware specifications.
Lastly, in \Sref{sec:datasets}, we describe 
the datasets we used to train and evaluate \DeMR, and evaluation metrics to measure the performance.

\section{Analysis on the Effects of the Sub-keypoints}  \label{sec:subkeypoint}
In this section, we explain the different settings of the
number of sub-keypoints and discuss why we set the 10 sub-keypoints setting as our full model in training time.
Moreover, we show the effectiveness of sub-keypoints with heterogeneous mesh prediction result.
\subsection{Sub-keypoint Selection}
\begin{figure*}[h]
\centering
    \subfigure[]{\includegraphics[width=0.25\linewidth]{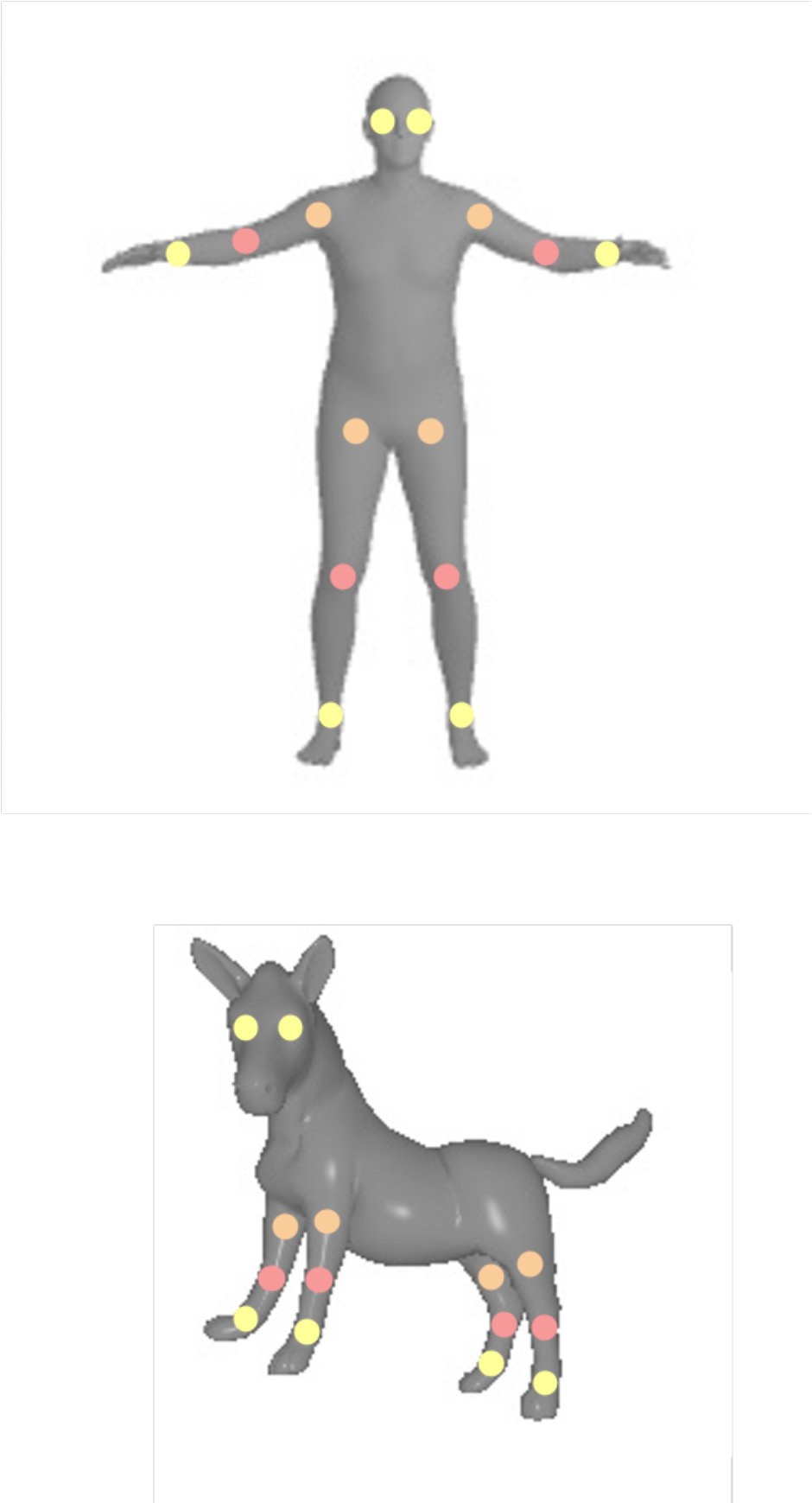}}
    \hspace{0.5cm}
    \subfigure[]{\includegraphics[width=0.25\linewidth]{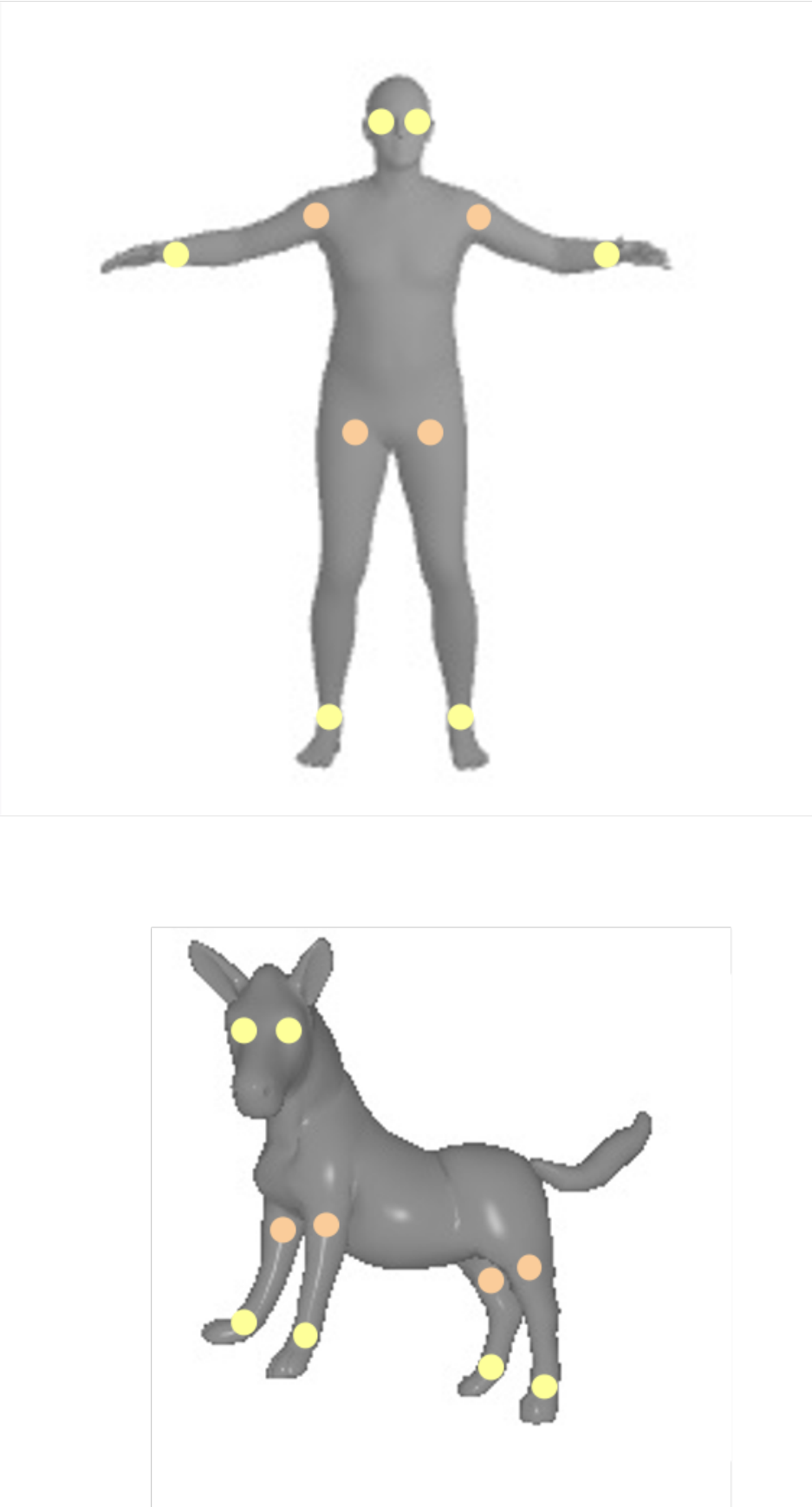}}
    \hspace{0.5cm}
    \subfigure[]{\includegraphics[width=0.25\linewidth]{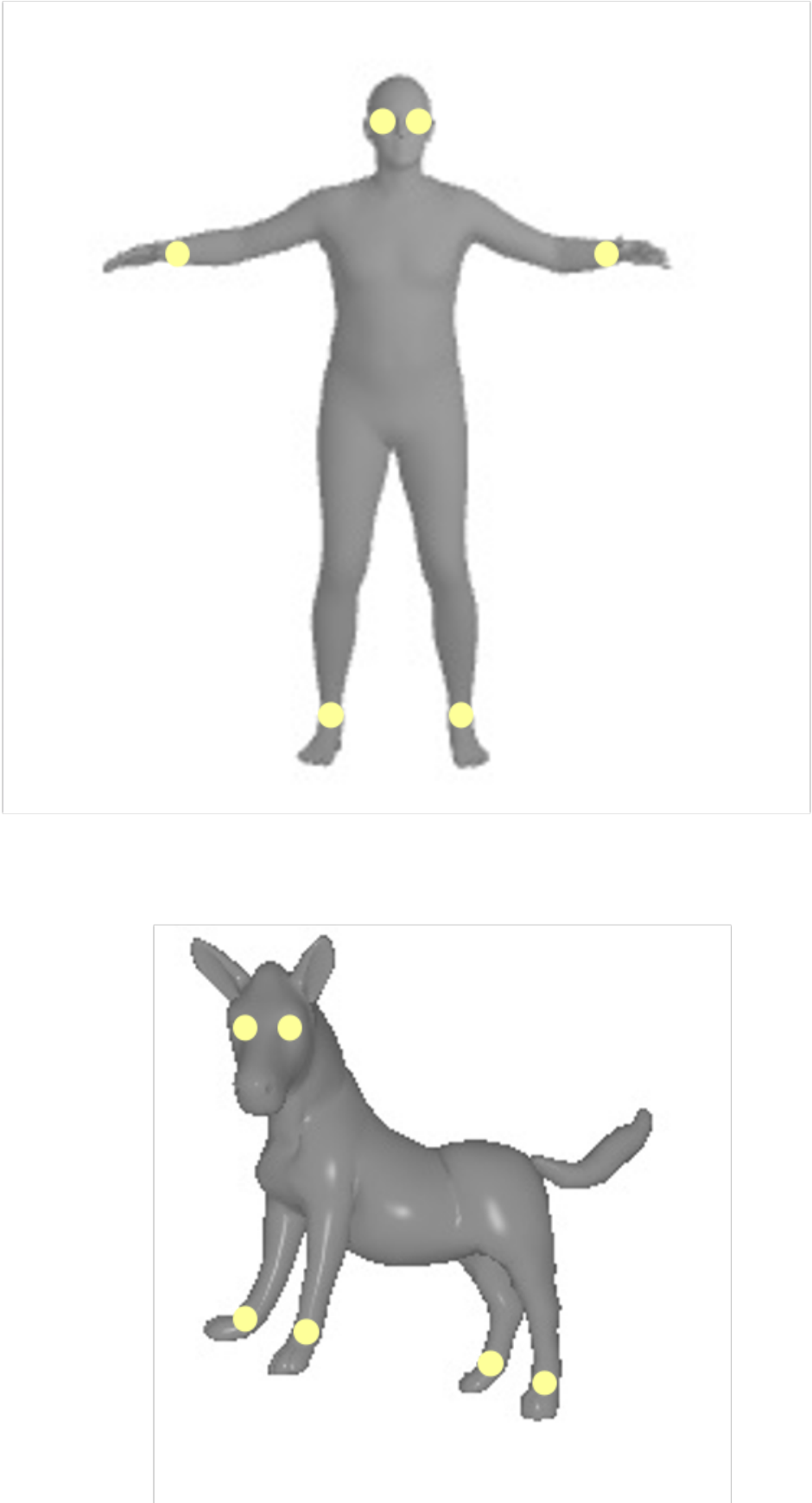}}
\vspace{3mm}
\caption{\textbf{Sub-keypoint settings}. (a) 15 sub-keypoints setting, including \{\emph{Eyes, Nose, TL, ML, BL}\}. (b) 10 sub-keypoints setting (Ours), including \{\emph{Eyes, Nose, TL, BL}\}. (c) 6 sub-keypoints setting, including \{\emph{Eyes, BL}\}. Best viewed in color.}
\vspace{-6mm}
\label{fig:supp_subkpt}
\end{figure*}

The concept of sub-keypoints that we propose in the main manuscript is defined as a subset of full body keypoints using morphological similarity between humans and animals. 
Since most of the semantic correspondences are found on their specific body parts, \eg, \{left arm: left front leg\}, we use \emph{Eyes, Nose, Top-of-Limb (TL), Middle-of-Limb (ML)}, and \emph{Bottom-of-Limb (BL)} as sub-keypoints for both humans and animals.
The main concern is that the number of sub-keypoints can be interpreted as the amount of morphological similarity that we consider in training time. 
Therefore, we set different settings of sub-keypoints (see \Fref{fig:supp_subkpt}) and analyze them.

For the 15 sub-keypoints setting, we choose the maximum number of corresponding keypoints, including \emph{Eyes, Nose, TL, ML,} and \emph{BL}. 
Even though we fully consider available sub-keypoints, fitting all of them rather disturbs training because humans and animals have different rotation angle range on their legs.
%
Since animals can bend their legs in the opposite direction compared to humans so that poses taken by animals can be different with that of humans.
Therefore, we exclude \emph{Middle-of-Limb} for 10 sub-keypoints setting according to the priority order in morphological similarity.
The setting without \emph{ML} can more accurately represent morphological similarity between humans and animals and also show best performance on animal reconstruction.
On the other hand, the setting for 6 sub-keypoints considers only \emph{Eyes} and \emph{BL} that is the minimal option of selecting keypoints. This minimum setting is not sufficient for the model to learn proper similarity.
Based on these observations, we found 10 sub-keypoints setting is the most appropriate to embody morphological similarity among heterogeneous classes. In addition, developing the method to automatically select the proper sub-keypoint will be a promising future research direction.
\subsection{Effectiveness of Sub-keypoint}
\vspace{-2mm}
\begin{wrapfigure}{R}{0.45\textwidth}
\centering
    \includegraphics[width=\linewidth]{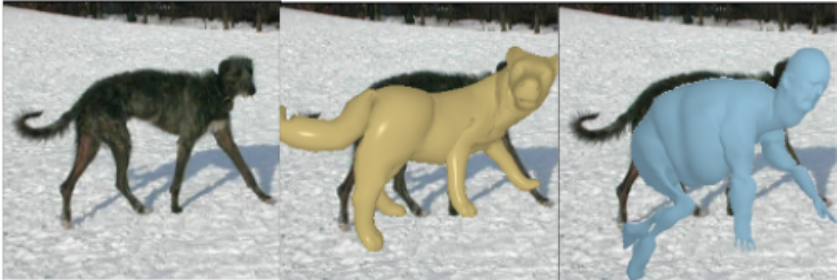}
\vspace{3mm}
\caption{{\textbf{Heterogeneous mesh prediction in test time}. Similarly-posed meshes for both human and animal show the effect of sub-keypoint.}}
\vspace{-4.5mm}
    \label{fig:supp_hetero}
\end{wrapfigure}
In the main manuscript, we discuss the effectiveness of using sub-keypoint in training time. The concept of sub-keypoint not only models
the morphological similarity between heterogeneous classes but also relaxes the disjoint multi-task learning problem into a jointly trainable multi-task setting. 
%
By using sub-keypoint, both human and animal keypoints are supervised by their counterparts, and the results of supervision are shown in \Fref{fig:supp_hetero}. As an animal image is taken as the input of the model, a human mesh who is imitating the animal can be predicted by guidance of morphological similarity. 
We observe that the introduction of sub-keypoint has a great influence on training the model, although the sub-keypoint loss accounts small percentage of a total loss; thus, such weak supervision is enough for \DeMR\ to learn morphological similarity. 
The demo video for the results of heterogeneous mesh prediction on BADJA dataset is provided as a supplementary video.


\section{Additional
Quantitative Results 
}   \label{sec:quan_eval}
In this section, we report our model's quantitative performance when trained with a larger model in terms of the capacity. In detail, we additionally increased the number of the parameters in the \emph{deformable body multi-task branch} 
by adding one more FC layer.
Since the SMAL mesh regression branch should cover three animal classes, increasing the branch's capacity would be helpful. 
\begin{table*}[t]
    \centering
    \resizebox{\columnwidth}{!}{%
    \begin{tabular}{l c c c c c c c}
        \toprule
        & \multicolumn{2}{c}{} & \multicolumn{1}{c}{MPJPE [mm] $\downarrow$} & \multicolumn{2}{c}{PA-MPJPE [mm] $\downarrow$} & \multicolumn{2}{c}{PCK [\%] $\uparrow$} \\\cmidrule(lr){2-3} \cmidrule(lr){4-4} \cmidrule(lr){5-6} \cmidrule(lr){7-8}
        \multirow{2}{*}{Models}& \multirow{2}{*}{\# of Params.} & \multirow{2}{*}{Trained Object Classes} &  MPI-INF-3DHP & Human3.6M-P2 & 3DPW & Stanford Extra & Animal Pose \\ 
        
        & &  & (Human) & (Human) & (Human) & (Dog) & (Horse, Cow) \\

        \cmidrule{1-8}
        
        HMR~\cite{hmrKanazawa17} & 27M & Human & 169.50 & 66.50 & 81.30 & - & -\\
        WLDO~\cite{biggs2020wldo} & 95M & Dog & - & - & - & 78.8 & - \\
        
        \cmidrule{1-8}
        
        Ours (4cls, na\"ive) & \multirow{3}{*}{27M+5M} & \multirow{3}{*}{Human, Dog, Horse, Cow} & 154.57 & 81.01 & 70.74 & 72.79 & 51.14 \\
        \quad \emph{+ Sub} & & & 155.34 & 82.42 & 69.94 & 72.90 & 52.72 \\
        \quad \emph{+ Sub + CSBN (full)} &  & &\underline{\textbf{140.76}} & \underline{\textbf{79.70}} & \underline{\textbf{69.85}} & 73.23 & 50.09 \\
        
        \cmidrule{1-8}
        Ours (4cls, 3FC) & \underline{\textbf{27M+6M}} & Human, Dog, Horse, Cow & - & - & - & \underline{\textbf{73.97}} & \underline{\textbf{54.56}} \\
        
        
        \bottomrule
    \end{tabular}
    }
    \vspace{2mm}
    \caption{Quantitative evaluation results of \DeMR\ on various human/animal datasets. The last row shows the reconstruction performance of our method, when trained with a larger model.}
    \label{table:supp}
    \vspace{-3mm}
\end{table*}

We trained the model with increased parameter size (27M+6M) for 300 epochs and evaluated PCK of both Stanford Extra dataset~\cite{biggs2020wldo,Khosla2012NovelDF} for dog and Animal Pose for horse and cow (see \Tref{table:supp}). 
\DeMR\ showed improved performances for both datasets. Especially, the performance for Animal Pose increased about 3.5\% compared to the main paper's full model. 
In this work, we focused on building a unified model with novel ideas such as sub-keypoint rather than achieving just a higher performance. However, although \DeMR\ is effective and compact in terms of the parameter numbers, \Tref{table:supp} tells one can achieve higher numerical performance when compromising with a larger capacity. 
%
\section{Additional
Qualitative Results
}    \label{sec:qual_eval}
\begin{figure*}[h!]
\centering
    \includegraphics[width=0.95\linewidth]{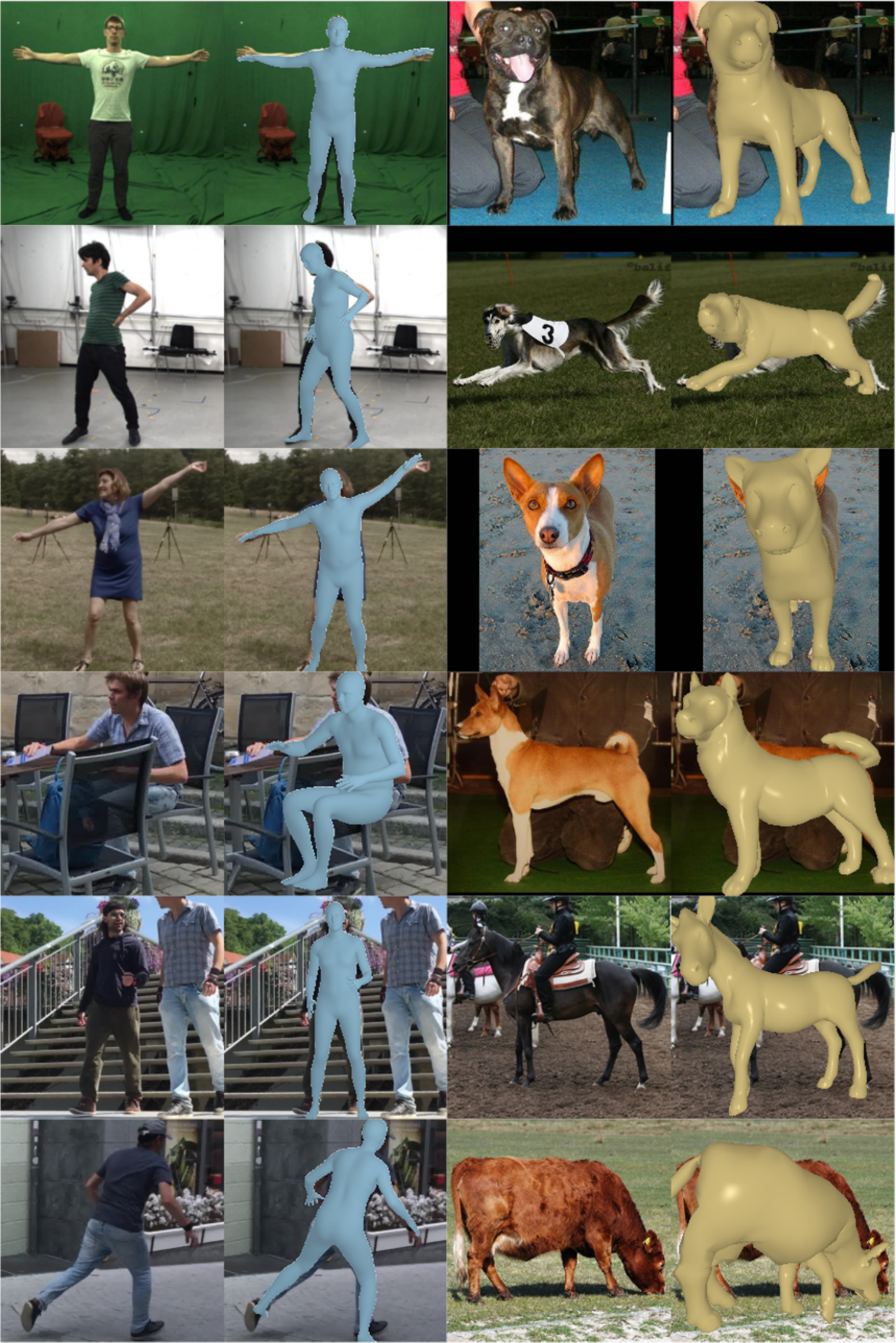}
\vspace{4mm}
\caption{\textbf{Additional Qualitative Results}.The first two columns show the input images of humans and SMPL mesh prediction results. The last two columns show the input images of animals and SMAL mesh prediction results.}
\vspace{-3mm}
    \label{fig:supp_qual}
\end{figure*}

In this section, we provide additional
qualitative evaluation of our model. Mainly, we compare \DeMR's 3D reconstruction performance with the animal uni-modal competing method, WLDO~\cite{biggs2020wldo}. 
For this qualitative comparison,
we used BADJA dataset~\cite{biggs2018creatures}, which is a benchmark for animals that contains multiple animal classes in it. Then, we provide additional mesh prediction results for various human and animal datasets~\cite{mono-3dhp2017,vonMarcard2018,biggs2020wldo,Cao2019CrossDomainAF} (see \Fref{fig:supp_qual}).

\begin{figure*}[h]
\centering
    \includegraphics[width=\linewidth]{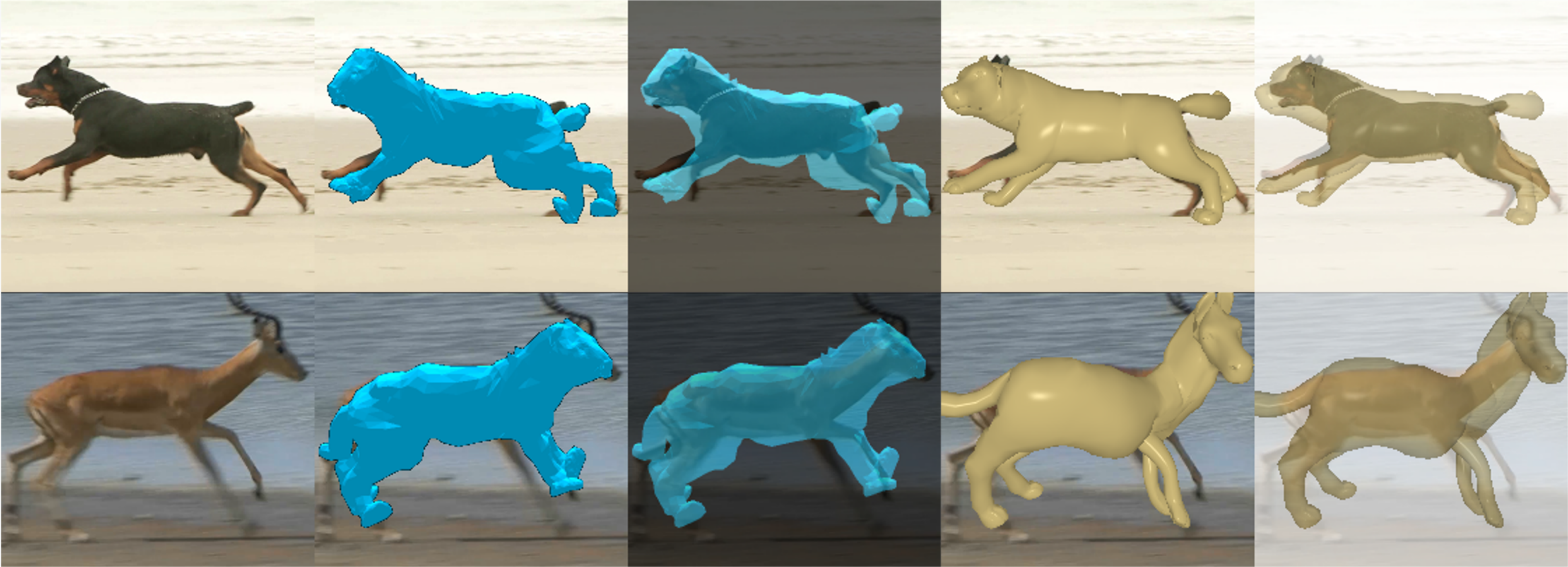}
\vspace{0.5mm}
\caption{\textbf{Comparison with WLDO}. Each column shows the input image, WLDO's mesh prediction~(blue), WLDO's mesh silhouette, \DeMR's mesh prediction~(yellow), \DeMR's mesh silhouette for BADJA: rs\_dog sequence (Top row), and for BADJA: impala0 sequence (Bottom row).}
\vspace{-4.5mm}
    \label{fig:supp_comp}
\end{figure*}

\para{BADJA: rs\_dog (seen class)} First, we compare \DeMR's 3D reconstruction performance for dog images with WLDO. 
Although WLDO is an expert on dog mesh reconstruction and \DeMR\ is made to cover multiple classes, \DeMR\ shows qualitatively better mesh predictions for dog images (see \Fref{fig:supp_comp}-(Top)).
WLDO shows bumpy mesh surfaces and not dog-like shaped mesh, despite its dog-specific shape regularizations. 
On the other hand, our method shows a more realistic and smooth mesh shape. 
In addition, the lower right leg of the predicted mesh from WLDO is in front of the lower left leg, which is inaccurate (see original image). 
However, \DeMR\ shows more accurate pose reconstruction than WLDO. 
Finally, \DeMR\ shows a better mesh silhouette for the dog. 
Silhouette for lower legs or the neck is more accurate in our method than WLDO.
This result shows the \DeMR's effectiveness in terms of the number of model parameters achieved by the introduction of sub-keypoint.

\para{BADJA: impala (unseen class)} In addition, we test both \DeMR\ and WLDO whether they can deal with unseen classes during training time, \eg, impala; thus, we test their generalizability.
Since WLDO is not trained with other animal classes, it ends up predicting randomly-shaped mesh, although it tried to estimate approximate pose for the impala. 
However, thanks to the \emph{class-selective shape prior loss} proposed in our method, although \DeMR\ have not seen the class, impala, during training time, our method can reconstruct more impala-like shaped mesh and pose just from a single image.
This result shows the powerful
generalization 
of our method, while proving the effectiveness of the proposed class-specific shape prior loss.

\vspace{-3mm}
\section{Training Details}  \label{sec:traning_details}
We provide training details, including hyper-parameters, training hardware specifications, and training time. 
\DeMR\ requires four main loss functions in order to learn morphological similarity across heterogeneous classes. 
Recall that the total training objective is given as follows:\vspace{-1.5mm}
\begin{equation}
\label{eq:tot_loss}
     \small{L =\alpha (\lambda_{\scriptscriptstyle{f}}^{h}L_{\scriptscriptstyle{full} }^{h}{+}\lambda_{\scriptscriptstyle{s}}L_{\scriptscriptstyle{sub}}^{h})+(1{-}\alpha)(\lambda_{\scriptscriptstyle{f}}^{a}L_{\scriptscriptstyle{full}}^{a}{+}\lambda_{\scriptscriptstyle{s}}L_{\scriptscriptstyle{sub}}^{a})} 
    + \lambda_{\scriptscriptstyle{sil}}L_{\scriptscriptstyle{sil}} + \lambda_{\scriptscriptstyle{sh}}^{a}L_{\scriptscriptstyle{sh}}^{a},
\end{equation}
The weights for the losses in \Eref{eq:tot_loss} were set to $\lambda_{\scriptscriptstyle{f}}^{h}{=}15$, $\lambda_{\scriptscriptstyle{f}}^{a}{=}5$, $\lambda_{\scriptscriptstyle{s}}{=}0.1$, $\lambda_{\scriptscriptstyle{sil}}{=}0.0075$, $\lambda_{\scriptscriptstyle{sh}}^{a}{=}10^{-5}$, respectively. We additionally used pose prior losses for both humans and animals, shape prior loss for humans to further regularize the model to predict natural bodies. Those auxiliary losses were directly adopted from previous human and animal uni-modal models~\cite{kolotouros2019spin, biggs2020wldo}. Our model was trained using the AdamW optimizer with a batch size of 64, and the learning rate set to $5\times10^{-5}$. In these conditions, overall training required 200 epochs to converge and took about 40 hours with one NVIDIA TITAN RTX GPU.

\vspace{-3mm}

\section{Datasets and Evaluation Metrics}   \label{sec:datasets}
In this section, we explain datasets we used in test time. In addition, we introduce three different evaluation metrics to quantitatively measure our method's favorable performance on other competing methods.
\subsection{Datasets}
We use 2D and 3D keypoint annotated datasets in test time to evaluate our method's 3D reconstruction performance. The evaluation datasets we used are 
as follows:

\begin{wrapfigure}{r}{0.42\linewidth}
\centering
    \vspace{-3mm}
    \includegraphics[width=\linewidth]{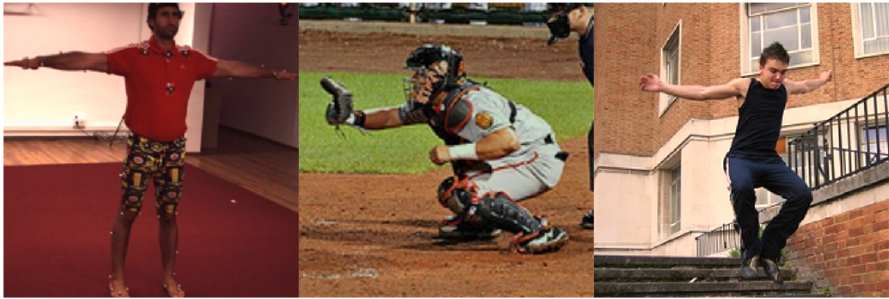}
    \vspace{-2mm}
    \caption{{\textbf{Dataset comparison.} 
    Human3.6M (left) shows distinct image characteristics, such as limited human poses in controlled indoor scenes, compared to the training datasets, COCO (middle) and LSP-extended (right), respectively. 
    }}
    \vspace{-6mm}
\label{fig:dataset_img}
\end{wrapfigure}

\para{HUMAN3.6M~\cite{h36m_pami}}
It is a large-scale video dataset with 3D human keypoint annotations captured in the indoor environment. We used `Protocol 2' of the dataset following the previous human competing method~\cite{hmrKanazawa17}.

\para{MPI-INF-3DHP~\cite{mono-3dhp2017}}
It is a dataset for 3D human body pose estimation captured mostly in indoor environments, with a small number of outdoor samples. We used the test splits for the evaluation.

\para{3DPW~\cite{vonMarcard2018}}
It is a human video dataset captured in outdoor settings. 3D annotations are obtained using video and IMU sensors. We use the test split to evaluate our model.

\para{StanfordExtra~\cite{biggs2020wldo,Khosla2012NovelDF}} 
It is a large-scale dataset for dogs with 2D keypoints and silhouette annotations. It covers 120 breeds of dogs with diverse shapes and poses. Along with keypoint location, binary segmentation masks are used to compute the silhouette loss and give a threshold to the PCK metric.

\para{Animal Pose Dataset~\cite{Cao2019CrossDomainAF}}
It is a 2D keypoint annotated animal dataset covering 5 kinds of animals: dog, cat, cow, horse, and sheep. We selected 2 classes, \ie, horse and cow, to train and evaluate the model. 

\subsection{Evaluation Metrics}
For the evaluation, three different metrics, \ie, \emph{PCK}, \emph{MPJPE}, \emph{PA-MPJPE} are used. The unit of PCK metric is in percentage~(\%) while the others are in millimeter~(mm).

\para{PCK}
Percentage of Correct Keypoint (PCK) is a metric that evaluates the quality of keypoint prediction. It is the ratio of the predicted keypoints that are within the threshold distance from the ground truth keypoints. The threshold is normalized by silhouette area; details can be found in~\cite{biggs2020wldo}. Since animal datasets do not contain 3D keypoint annotations, we measured PCK by re-projecting predicted 3D keypoints of animals into 2D. 

\para{MPJPE}
Mean per joint position error (MPJPE) is a metric that measures the 3D keypoint reconstruction error. It computes the mean of the Euclidean distances between the predicted keypoints and the ground truth keypoints. We use MPJPE to evaluate our model for the human datasets with 3D ground truth annotations.

\para{PA-MPJPE}
PA-MPJPE also measures the 3D keypoint reconstruction error, but after the rigid alignments of reconstructed keypoints to the ground truth keypoints using Procrustes Analysis~\cite{procrustes}. Thus, PA-MPJPE is effective when computing the 3D keypoint error regardless of global misalignments. We used PA-MPJPE to evaluate our model on the 3D human datasets.

\clearpage

\bibliography{egbib}